\documentclass[a4paper,fleqn]{cas-dc}

\usepackage[authoryear]{natbib}
\usepackage{csquotes}
\usepackage{amsmath,amssymb,amsfonts}
\usepackage{algorithmic}
\usepackage{textcomp}
\usepackage{bm}

\usepackage{multirow}
\usepackage{manyfoot}
\usepackage{booktabs}
\usepackage{listings}
\usepackage[nolist]{acronym}
\usepackage[english]{babel}
\usepackage{makecell}
\usepackage[]{hyperref}
\usepackage{wrapfig}
\usepackage{siunitx}
\usepackage[export]{adjustbox}
\usepackage{url}
\usepackage{nicefrac}

\newcommand{\reffig}[1]{Figure~\ref{#1}}
\newcommand{\refsec}[1]{Section~\ref{#1}}
\newcommand{\reftab}[1]{Table~\ref{#1}}
\newcommand{\refeq}[1]{\eqref{#1}}

\renewcommand{\vec}[1]{\bm{#1}}
\newcommand{\diag}[0]{\text{diag}}
\newcommand{\paperlink}{\url{https://mrs.fel.cvut.cz/papers/usv-state-estimation}}
\newcommand{\usveslinear}{\mbox{6DOF-L}}
\newcommand{\usvesnonlinear}{\mbox{6DOF-N}}
\newcommand{\sota}{SOTA}

\usepackage{tikz}
\definecolor{accessblue}{cmyk}{1, 0.3, 0, 0.2}
\definecolor{greycolor}{cmyk}{0,0,0,.8}
\usepackage{pgfplots}
\pgfplotsset{compat=1.14}
\usetikzlibrary{quotes, angles, backgrounds, arrows, automata, shapes, positioning, calc, through, spy, decorations.pathreplacing, decorations.markings, arrows.meta, automata, petri, intersections, pgfplots.fillbetween}
\pgfdeclarelayer{background}
\pgfdeclarelayer{foreground}
\pgfsetlayers{background, main, foreground}

\tikzset{
  state/.style={
    rectangle,
    draw=black, very thick,
    minimum height=1.0em,
    text centered,
  },
  smallstate/.style={
    rectangle,
    draw=black, very thick,
    minimum height=0.2em,
    text centered,
  },
  final_state/.style={
    rectangle,
    rounded corners,
    draw=black, very thick,
    minimum height=2em,
    text centered,
  },
  initial_state/.style={
    rectangle,
    double=white,
    double distance=1pt,
    inner sep=2pt,
    draw=black, very thick,
    minimum height=2em,
    text centered,
  },
  point/.style={
    circle,
    inner sep=0pt,
    minimum size=3pt,
    fill=red
  },
  adder/.style={
    circle,
    inner sep=2pt,
    minimum size=0.3in,
    draw=black, very thick,
    text centered
  },
  state_gray/.style={
    rectangle,
    draw=black, very thick,
    fill=gray!40,
    minimum height=1.0em,
    text centered,
    inner sep=0,
  },
  state_white/.style={
    rectangle,
    draw=black, very thick,
    fill=white,
    minimum height=1.0em,
    text centered,
    text=black,
    inner sep=0,
  },
  state_green/.style={
    rectangle,
    draw=black, very thick,
    fill=green!50,
    minimum height=1.0em,
    text centered,
    text=black,
    inner sep=0,
  },
  state_red/.style={
    rectangle,
    draw=black, very thick,
    fill=red!70,
    minimum height=1.0em,
    text centered,
    text=black,
    inner sep=0,
  },
  state_blue/.style={
    rectangle,
    draw=black, very thick,
    fill=blue!40,
    minimum height=1.0em,
    text centered,
    text=black,
    inner sep=0,
  },
  final_state/.style={
    rectangle,
    rounded corners,
    draw=black, very thick,
    minimum height=2em,
    text centered,
  },
  initial_state/.style={
    rectangle,
    double=white,
    double distance=1pt,
    inner sep=2pt,
    draw=black, very thick,
    minimum height=2em,
    text centered,
  },
  point/.style={
    circle,
    inner sep=0pt,
    minimum size=3pt,
    fill=red
  },
}

\tikzset{new spy style/.style={spy scope={
  magnification=5,
  size=1.25cm,
  connect spies,
  every spy on node/.style={
    rectangle,
    draw,
  },
  every spy in node/.style={
    draw,
    rectangle,
    fill=white
  }
  }
  }
  }

\def\tsc#1{\csdef{#1}{\textsc{\lowercase{#1}}\xspace}}
\tsc{WGM}
\tsc{QE}

\newcommand{\PREPRINTYEAR}{2024}
\newcommand{\PUBLISHEDIN}{Ocean Engineering}
\newcommand{\DOI}{10.1016/j.oceaneng.2025.120606} 

\usepackage[placement=top,vshift=-7]{background}
\SetBgScale{1.0}
\SetBgContents{\PUBLISHEDIN. PREPRINT VERSION - DO NOT DISTRIBUTE. \href{https://doi.org/\DOI}{DOI \DOI}}
\SetBgColor{black}
\SetBgAngle{0}
\SetBgOpacity{1.0}

\begin{document}

\thispagestyle{empty}
\onecolumn
{
  \topskip0pt
  \vspace*{\fill}
  \centering
  \LARGE{%
    \copyright{} \PREPRINTYEAR~\PUBLISHEDIN\\\vspace{1cm}
    This manuscript version is made available under the CC-BY-NC-ND 4.0 license \url{https://creativecommons.org/licenses/by-nc-nd/4.0/}\vspace{1cm}\newline
    \large DOI: \href{https://doi.org/\DOI}{\DOI}
    \vspace*{\fill}}
    \vspace*{\fill}
}
\NoBgThispage
\twocolumn          	
\BgThispage

\let\WriteBookmarks\relax
\def\floatpagepagefraction{1}
\def\textpagefraction{.001}




\shorttitle{State estimation of marine vessels affected by waves by unmanned aerial vehicles}
\shortauthors{F. Novak et al.}

\title[mode = title]{State estimation of marine vessels affected by waves by unmanned aerial vehicles}  

\tnotemark[1,2]
\tnotetext[1]{This work has been supported by the Technology Innovation Institute - Sole Proprietorship LLC, UAE, under the Research Project Contract No. TII/ARRC/2055/2021, CTU grant no SGS23/177/OHK3/3T/13 and the Czech Science Foundation (GAČR) under research project No. 23-07517S.
}
\tnotetext[2]{Multimedia materials: \paperlink{}}

\author[1]{Filip Nov\'{a}k}[type=editor,
    auid=000,
    bioid=1,
    orcid=0000-0003-3826-5904]
\cormark[1] 
\ead{filip.novak@fel.cvut.cz} 
\ead[url]{https://mrs.fel.cvut.cz/filip-novak}
    
\author[1]{Tom\'{a}\v{s} B\'{a}\v{c}a}[type=editor,
    auid=000,
    bioid=1,
    orcid=0000-0001-9649-8277]
    
\author[1]{Ond\v{r}ej Proch\'{a}zka}[type=editor,
    auid=000,
    bioid=1,
    orcid=0009-0009-2224-750X]

\author[1]{Martin Saska}[type=editor,
    auid=000,
    bioid=1,
    orcid=0000-0001-7106-3816]

\address[1]{Department of Cybernetics, Faculty of Electrical
Engineering, Czech Technical University in Prague, Czech Republic}

\cortext[1]{Corresponding author} 

\begin{abstract}
A novel approach for robust state estimation of marine vessels in rough water is proposed in this paper to enable tight collaboration between Unmanned Aerial Vehicles (UAVs) and a marine vessel, such as cooperative landing or object manipulation, regardless of weather conditions.
Our study of marine vessel (in our case Unmanned Surface Vehicle (USV)) dynamics influenced by strong wave motion has resulted in a novel nonlinear mathematical USV model with 6 degrees of freedom (DOFs), which is required for precise USV state estimation and motion prediction.
The proposed state estimation and prediction approach fuses data from multiple sensors onboard the UAV and the USV to enable redundancy and robustness under varying weather conditions of real-world applications.
The proposed approach provides estimated states of the USV with 6 DOFs and predicts its future states to enable tight control of both vehicles on a receding control horizon.
The proposed approach was extensively tested in the realistic Gazebo simulator and successfully experimentally validated in many real-world experiments representing different application scenarios, including agile landing on an oscillating and moving USV.
A comparative study indicates that the proposed approach significantly surpassed the current state-of-the-art.
\end{abstract}








\maketitle

\section{INTRODUCTION}
Tight collaboration between multirotor \acp{UAV} and Manned and \acp{USV}, i.e. ships or boats \citep{Fossen2011} is required in numerous offshore as well as freshwater applications.
The \acp{USV}-\acp{UAV} team is beneficial in searching for and removing garbage from water \citep{usv_uav_water_pollution}, providing assistance in the aftermath of disasters~\citep{usv_uav_hurricane_wilma, usv_uav_hurricane_wilma2}, transporting materials and objects from one place to another \citep{Steenken2001}, cooperation with rescue services \citep{usv_uav_rescue_system}, gathering environmental data and water quality monitoring \citep{usv_uav_measurement}.

\begin{figure}[!t]
  \centering
  \includegraphics[width=\linewidth]{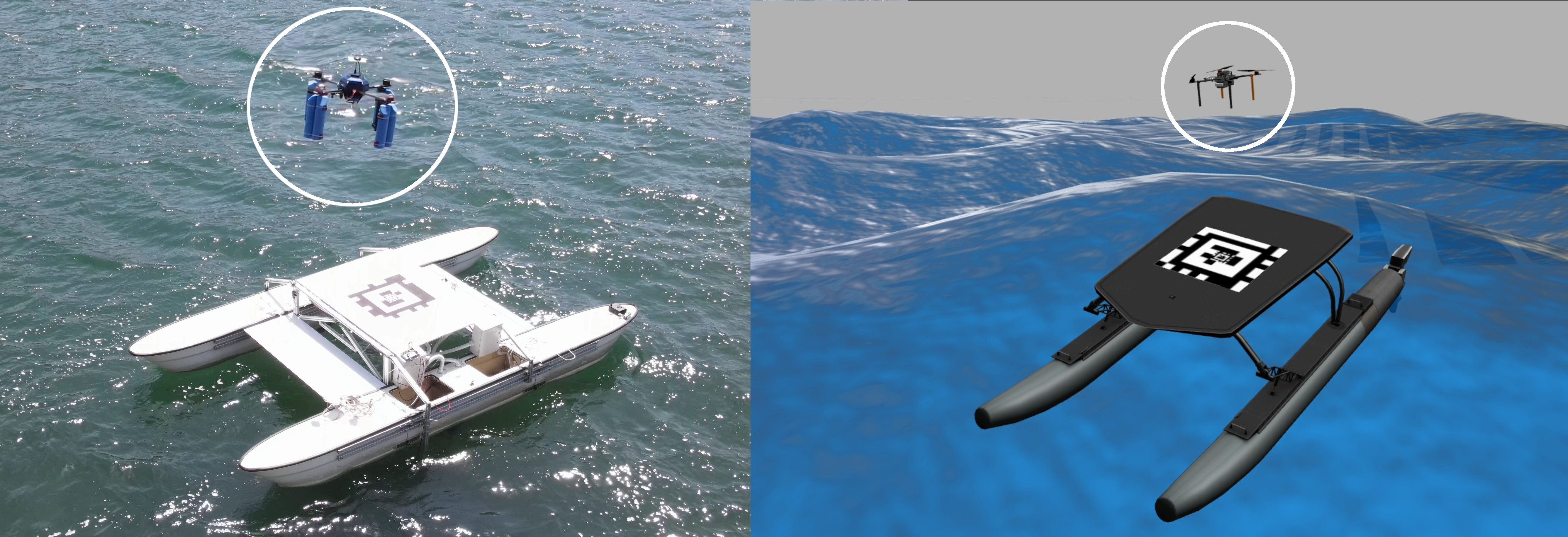}
  \caption{UAV used in the presented research cooperating together with a USV equipped with a landing platform.}
  \label{fig:usv_uav}
\end{figure}

An important aspect of any tight multi-robot collaboration is mutual localization among the teammates \citep{uvdd2}.
Relative states between \acp{UAV} and \acp{USV}, which consist of position, orientation, velocity, and angular velocity, are required in most of complex multi-robot missions \citep{uav_usv_ladning_tor}.
Using this information, the \ac{UAV} can precisely follow the \ac{USV} \citep{prochazka2024ModelPredictiveControlbased, uav_follow_usv}, explore different parts of the environment \citep{usv_uav_hurricane_wilma2}, or land on the \ac{USV} \citep{prochazka2024ModelPredictiveControlbased, uav_usv_landing5, zhang2024RobustAutonomousLandinga, uav_usv_landing_parakh, uav_usv_landing3, kwak2022AutonomousUAVTarget, Xu2020_vision, uav_usv_landing2}.
The landing task was selected as a key study of the approach presented in this paper due to its sensitivity to precise data. 
In order to land on the \ac{USV}, the \ac{UAV} requires accurate localization and state estimation of the landing platform placed onboard the \ac{USV} \citep{prochazka2024ModelPredictiveControlbased} and motion prediction of the \ac{USV} in case of the rough sea surface (\reffig{fig:usv_uav}).

The ability of the \ac{UAV} to land on the \ac{USV} in all weather and sea conditions is required for battery recharging, as \ac{UAV} flight time is greatly limited by battery capacity \citep{uav_usv_design_platform}. 
Other examples include unloading cargo carried by the \ac{UAV} \citep{uav_dock_stations} and transporting the \ac{UAV} to the area of its operation \citep{usv_uav_platform}.
The \ac{USV} can also provide a safe docking spot for the \ac{UAV} in the event of inclement weather \citep{uav_usv_design_platform}.
A safe landing, while avoiding critical situations leading to damage to the aerial vehicle as shown in \reffig{fig:uav_usv_landing_motivation}, requires designing a robust approach for estimating and predicting \ac{USV} states with 6 \acp{DOF}, as well as multi-sensor perception for various weather conditions, as presented in this paper.

\section{RELATED WORKS}
\label{sec:related_works}
The approaches to state estimation can be separated into two groups.
The first group consists of systems relying on \ac{GNSS} (e.g., \ac{GPS}) and \ac{IMU} to estimate the states \citep{gps_ship,gps_usv, gps_nav}.
To obtain the \ac{USV} states relative to the \ac{UAV} states, the estimated states of the \ac{USV} must be sent to the \ac{UAV} via a reliable communication link with sufficient rate and bandwidth.
However, the required communication link is challenging to achieve in real-world deployment \citep{TRAN201967}.
Moreover, the precision of \ac{GNSS} is not sufficient in order of magnitude for landing \ac{UAV} into a recharging box, which is often required \citep{GPSaccuracy}.

State estimation methods in the second group use vision-based relative localization systems to provide information about desired targets \citep{krogius2019iros, uvdd1, uvdd2, usv_uav_navigation, wang2016iros, olson2011tags}.
The concept of these systems enables them to be put onboard \acp{UAV} that use them to estimate \ac{USV} states without the need for any communication link.
However, such relative localization systems require a target in a detectable position.
Usually, the system must be close to the target, and the vision sensor, e.g., the camera, must see the target in its frame to provide measurements.
Further, the performance and reliability of these systems depend on weather and lighting conditions \citep{krogius2019iros, wang2016iros, olson2011tags}, and the estimation of target tilt using only camera images is inaccurate \citep{uvdd2, uvdd1, usv_uav_navigation, olson2011tags}.

In all approaches discussed above, the gained sensor data need to be processed in real-time to estimate the desired states for a fast control loop.
Most of the approaches address real-time state estimation using the Kalman filter \citep{ukf_ras, Malyuta2019, usv_navigation_ukf, KalmanFossen, Julier1997728, kalman1960new}.
The Kalman filter fuses measurements from different sensors to obtain accurate state estimation of a given system \citep{ribeiro2004kalman, Haykin20011001}, and further requires a mathematical model of the system to estimate its states.

The full 6 \ac{DOF} nonlinear mathematical models of \ac{USV} are presented in \citep{Fossen2011, Fossen2002, Fossen1994}.
These works also present linearization of the nonlinear \ac{USV} model and discuss the methods for wave motion filtering.
However, the estimation of the \ac{USV} states with 6 \acp{DOF} in waves is not presented.
The approach \citep{KalmanFossen} uses a simplified 3 \ac{DOF} linear model based on \citep{Fossen2011, Fossen2002, Fossen1994} for positioning and heading control.
The 3 \ac{DOF} mathematical \ac{USV} models consisting of planar position and heading are also considered in \citep{Wang2021_observer, Xu2020_vision, Wirtensohn2016518, tomera2012nonlinear}.
As discussed later, the 3 \ac{DOF} state estimation is not sufficient in harsh conditions as the heave motion and roll and pitch angles are not observable by these approaches.
Instead of relying on rigorous mathematical models, probabilistic models and reinforcement learning techniques can be employed \citep{xia2024EstimatingLyapunovRegion, cui2022FilteredProbabilisticModel, cui2021AutonomousBoatDriving}.
However, these methods typically require extensive learning data, numerous training trials, or iterative processes.
Additionally, they are generally more computationally demanding compared to classical methods.

\begin{figure}[!tb]
  \centering
  \input{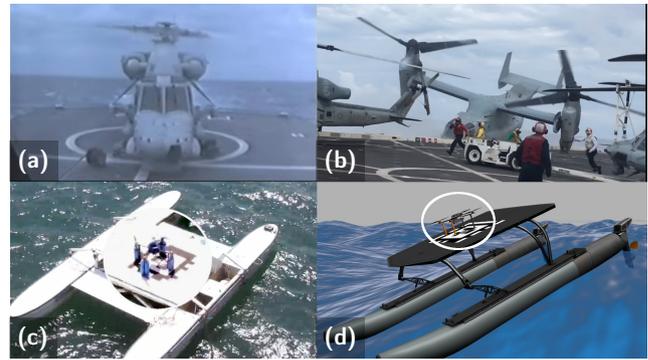}
  \caption{Unsuccessful landing on a marine vessel without the landing pad motion detection and prediction (mainly the heave motion was difficult to observe), shown in (a) and (b), serves as motivation to create a robust USV state estimation system in 6 DOFs to ensure safe autonomous landing in rough water and different real-world conditions as shown in snapshots from our system deployment (c) and (d).}
  \label{fig:uav_usv_landing_motivation}
\end{figure}

\begin{table*}
\caption{A comparison of the available USV state estimation approaches related to this work.}
\centering
\begin{tabular}{lcccccc}
\hline
\\[-0.75em]
Work & \makecell{Designed\\for UAVs} & \makecell{Number\\of \acsp{DOF}} & \makecell{Wave dynamics\\in USV model} & \makecell{Predict future\\ states including\\ wave motions} & \makecell{Ability to fuse\\ sensor data from\\ UAV and USV} & \makecell{Real-world\\deployment}\\
\\[-0.75em]
\hline
\\[-0.75em]
\cite{wang2021_dual_ukf}  & No & 3 & No & No & No & No\\
\cite{tomera2012nonlinear} & No & 3 & Yes & No & No & Yes\\
\cite{KalmanFossen} & No & 3 & Yes & No & No & No\\
\cite{Wirtensohn2016518} & No & 3 & Yes & No & No & Yes\\
\cite{Xu2020_vision} & Yes & 3 & No & No & No & Yes\\
\cite{Abujoub2018_landing} & Yes & 5 & No & Yes & No & No\\
\cite{uav_usv_landing2} & Yes & 6 & No & No & No & No\\
\textbf{This work} & \textbf{Yes} & \textbf{6} & \textbf{Yes} & \textbf{Yes} & \textbf{Yes} & \textbf{Yes}\\
\\[-0.75em]
\hline
\end{tabular}
\label{table:ComparisonRelatedWorks}
\end{table*}

From the available literature we selected the most relevant papers to our approach.
\reftab{table:ComparisonRelatedWorks} summarizes the comparison of these state-of-the-art works with the approach proposed in this paper.
Most of the works use simplified \ac{USV} state vectors representing models with 3 \acp{DOF} (planar translation and heading) \citep{wang2021_dual_ukf, Xu2020_vision, Wirtensohn2016518, tomera2012nonlinear, KalmanFossen}.
The system presented in \citep{Abujoub2018_landing} estimates the \ac{USV} with 5 \acp{DOF}.
The approach presented in \citep{uav_usv_landing2} estimates the \ac{USV} with 6 \acp{DOF}.
However, the wave motion is not considered in \citep{uav_usv_landing2}, which could lead to critical landing failures (see \reffig{fig:uav_usv_landing_motivation} (a) and (b)) and restricts using the work in real-world conditions.
Accounting for wave motion is crucial because it directly impacts the accuracy of the estimated amplitude and frequency of the \ac{USV}’s movement.
Without this consideration, the \ac{UAV} may attempt to land too early, colliding with the landing platform at excessive speed, or miss the platform entirely due to unanticipated downward motion.
Additionally, inaccurate estimation of wave effects, particularly during platform tilting as the \ac{UAV} attempts to land, may lead to uneven touchdowns and result in hazardous interactions with the \ac{USV}'s deck.
Such errors can result in critical landing failures and make the system unsuitable for real-world applications, particularly with smaller vessels heavily affected by wave dynamics.
In the available literature, the wave dynamics in \ac{USV} model during estimation of the \ac{USV} states is considered only with lower \acp{DOF} \citep{Wirtensohn2016518, tomera2012nonlinear, KalmanFossen}, which simplifies the problem and may lead to failures due to unobservable oscillation and heave motion (see \reffig{fig:uav_usv_landing_motivation}).
Our work overperforms the available literature by considering the most complex situation with a state vector with a full 6 \acp{DOF}, where wave motion is taken into account for all \ac{USV} states.

Regarding the ability to predict future \ac{USV} states, the approach in \citep{Abujoub2018_landing} is able to compute such a prediction to exhibit periodic behavior resembling wave motions.
However, only 2 \acp{DOF} (roll and pitch angles) are predicted, which is insufficient for tight cooperation and landing.
The approach \citep{Abujoub2018_landing} assumes observing waves in filtered raw sensor data as periodic motion. 
Then the main components of these periodic signals are identified using the Fast Fourier Transform and used to predict future roll and pitch angles.
Data are obtained from 3 simulated single pulse \ac{LiDAR} devices assuming that laser beam from each \ac{LiDAR} device hits the desired area on the landing board, which can be very problematic in a real-world deployment.
The work \citep{Abujoub2018_landing} was not verified in real-world conditions, and the considered assumptions on sensors uncertainty are not realistic based on our experimental analyses.
The works \citep{Xu2020_vision, Wirtensohn2016518, tomera2012nonlinear} deployed in the real world use only a \ac{USV} state vector with 3 \acp{DOF} and do not predict future \ac{USV} states.

\subsection{Contributions}
This paper proposes a novel solution for the robust 6 \ac{DOF} state estimation and prediction of a moving \ac{USV} on a rough water surface.
The primary contributions of the presented method, extending beyond the existing literature, include the development of a novel nonlinear 6 \ac{DOF} USV mathematical model incorporating wave dynamics.
Additionally, a linear version of the presented nonlinear model is derived.
Furthermore, the method enables the prediction of future \ac{USV} states with 6 \acp{DOF}.
These contributions have been rigorously validated through numerous simulations and real-world experiments.
The overall performance has been compared with state-of-the-art methods.

\subsection{Problem statement}
This paper tackles the problem of accurate \ac{USV} state estimation with 6 \acp{DOF} by a \ac{UAV} on a rough water surface and prediction of the future \ac{USV} states.
This enables the planning of UAV trajectories to be both optimal and efficient, ensuring reliable coordination with the USV and successful landing onboard the USV under harsh conditions, which are our target scenarios.
One of the addressed sub-problems is creating a 6 \ac{DOF} mathematical model of the \ac{USV} moving on a rough water surface, which is used for estimation and prediction.
The proposed estimation approach is designed for a real-world environment, where limits of individual global and onboard sensors are considered to ensure the robustness of the entire solution in real-world conditions.

It is assumed that the \ac{UAV} is equipped by \ac{GNSS} receiver, flight controller with \ac{IMU}, and down-looking cameras to utilize global and onboard localization systems, whose measurements are processed by the onboard CPU.
The \ac{USV} carries \ac{GNSS} receiver, \ac{IMU}, and pattern detectable from \ac{UAV} onboard sensors.
Moreover, a communication link between the \ac{UAV} and the \ac{USV} is required for transferring onboard sensor data from the \ac{USV} to the \ac{UAV}, at least on 1~Hz.
However, the designed system is able to work in the event of an unavailable communication link by using just the \ac{UAV} onboard sensors.
It is further assumed that the \ac{USV} is moving at a speed where the \ac{UAV} is able to follow it --- based on our experimental evaluation, we require the \ac{USV} speed to be less than 90\% of the maximal \ac{UAV} speed.

\section{MATHEMATICAL USV MODEL}\label{sec:usv_models}
The full nonlinear \ac{USV} model is used to fuse the data of multiple sensors and additionally predict future states of the \ac{USV}.
The approach proposed in this paper focuses on state estimation and prediction of the \ac{USV} moving on wavy water surfaces with a particular interest in parameters required for precise landing in rough conditions.
Therefore, the regularly used \ac{USV} model needs to be extended by wave dynamics to obtain sufficient estimation and prediction accuracy (\refsec{sec:waves_model}).

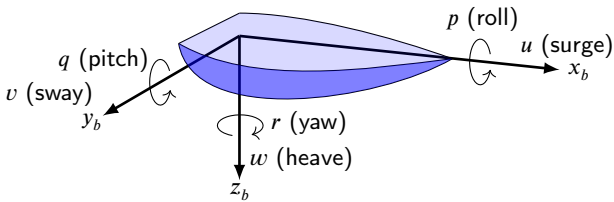
\begin{figure}[!b]
  \centering
  \usetikzlibrary{intersections}
\begin{tikzpicture}[scale=0.4]
    \draw[name path=Back] (0,9) -- (2,10);
    \draw[name path=Bottom] (0,9) to [out=-80,in=210,looseness=0.8] (9,8.5);
    \draw[name path=Right] (0,9) to [out=-30,in=187,looseness=0.8] (9,8.5);
    \draw[name path=Left] (2,10) to [out=0,in=160,looseness=0.75] (9,8.5);
    \draw[name path=a,line width=1pt, -latex] (2,9.25) -- (12.53,8.1385)
                            node[pos=1.06] {$x_b$}
                            node[above left, pos=1.2, yshift=3.25] {$u$ (surge)};
    \draw[name path=b,line width=1pt, -latex] (2,9.25) -- (-2.5,6.625) node[pos=1.08] {$y_b$} node[above left, pos=1] {$v$ (sway)};
    \draw[name path=c,line width=1pt, -latex] (2,9.25) -- (2,4.5) node[pos=1.08] {$z_b$} node[above right, pos=1] {$w$ (heave)};
    \tikzfillbetween[of=Bottom and Right]{blue, opacity=0.5};
    \tikzfillbetween[of=Left and Right]{blue, opacity=0.15};
    \draw[->] (10.25,8.65) arc (20:340:0.35 and 0.75) node[above, pos=1.3] {$p$ (roll)};
    \draw[->] (-0.3,8.1) arc (30:340:0.35 and 0.75) node[above, left, pos=1.4] {$q$ (pitch)};
    \draw[<-] (2.4,6) arc (-60:250:0.75 and 0.35) node[right, pos=1.4] {$r$ (yaw)};

    \draw[name path=xg,line width=1pt, -latex] (10.53,4.5) -- (13.03,4.025)
                            node[pos=1.13] {$x_g$};
    \draw[name path=yg,line width=1pt, -latex] (10.53,4.5) -- (8.23,3.575) 
                            node[pos=1.13] {$y_g$};
    \draw[name path=zg,line width=1pt, -latex] (10.53,4.5) -- (10.53,2) 
                            node[pos=1.13] {$z_g$};

    \draw[name path=rg,line width=1pt, -latex] (10.53,4.5) -- (2,9.25); 
    \node[] at (8.5,6.25) {$\vec{\eta}$};
    \node[] at (0,11.25) {Body-fixed coordinate frame};
    \node[] at (5,1.75) {Global coordinate frame};

\end{tikzpicture}
  \caption{Motion of the USV with 6 DOFs.}
  \label{fig:usv_body_states}
\end{figure}

The \ac{USV} is modeled as a rigid body with 6 \acp{DOF} of translation and rotation in 3D (see \reffig{fig:usv_body_states}).
The longitudinal motion of the \ac{USV} in direction of the $x_b$ axis is called $surge$.
The rotation around the $x_b$ axis is known as $roll$.
$Sway$ is lateral motion, or sideways motion, in the direction of the $y_b$ axis. 
The rotation corresponding to the $y_b$ axis is $pitch$.
The last motion, known as $heave$, takes place in a vertical direction with respect to the $z_b$ axis, and $yaw$ is the corresponding rotation.
\reftab{tab:usv_dof_notation} summarizes notation of \ac{USV} motion variables corresponding to the presented \acp{DOF}: position and orientation, their derivatives representing linear and angular velocities, and forces and torques. 

\begin{table}[!b]
  \caption{The notation of USV motion variables.}
  \centering
    \begin{tabular}{ccccc}
    \hline
    DOF & meaning & \makecell{Positions and\\ Euler angles} & \makecell{Linear and\\ angular\\ velocities} & \makecell{Forces\\ and\\ torques}\\ 
    \hline
    1   & surge & $x$ & $u$ & $\tau_x$\\ 
    2   & sway & $y$ & $v$ & $\tau_y$\\ 
    3   & heave & $z$ & $w$ & $\tau_z$\\ 
    4   & roll & $\phi$ & $p$ & $\tau_\phi$\\
    5   & pitch & $\theta$ & $q$ & $\tau_\theta$\\ 
    6   & yaw & $\psi$ & $r$ & $\tau_\psi$\\ 
    \hline
    \end{tabular}
  \label{tab:usv_dof_notation}
\end{table}

\subsection{Kinematics}
First, we define the kinematic transformation to describe the motion of the \ac{USV} in different coordinate frames.
During analysis of the \ac{USV} motion, two coordinate frames are defined as shown in \reffig{fig:usv_body_states}.
The first one is a body-fixed frame of the \ac{USV} that is put into the center of gravity of the \ac{USV}.
The second one is a global coordinate frame whose origin is fixed to the Earth's surface.
It is assumed that the position and orientation of the \ac{USV} are expressed in a global coordinate frame, while linear and angular velocities are described in a body-fixed coordinate frame.
According to the coordinate frames presented above, the following vectors describe the \ac{USV} motion
\begin{align}
    \vec{\eta} &= (\vec{p}^{\intercal},~\vec{\Theta}^{\intercal})^{\intercal} = (x,~y,~z,~\phi,~\theta,~\psi)^{\intercal},\label{eq:usv_global_states}\\
    \vec{\nu} &= (\vec{v}^{\intercal},~\vec{\omega}^{\intercal})^{\intercal} = (u,~v,~w,~p,~q,~r)^{\intercal},\label{eq:usv_body_states}\\ 
    \vec{\tau} &= (\vec{\tau}^{\intercal}_{\vec{p}},~\vec{\tau}^{\intercal}_{\vec{\Theta}})^{\intercal} = (\tau_x,~\tau_y,~\tau_z,~\tau_\phi,~\tau_\theta,~\tau_\psi)^{\intercal},
\end{align}
where $\vec{\eta}$ denotes position $\vec{p}=(x,~y,~z)^{\intercal}$ and orientation $\vec{\Theta}=(\phi,~\theta,~\psi)^{\intercal}$ in terms of intrinsic Euler angles \citep{diebel2006representing} in a global coordinate frame.
The vector $\vec{\nu}$ denotes linear velocity $\vec{v}=(u,~v,~w)^{\intercal}$ and angular velocity $\vec{\omega}=(p,~q,~r)^{\intercal}$ in a body-fixed coordinate frame.
The term $\vec{\tau}$ represents forces $\vec{\tau}_{\vec{p}}=(\tau_x,~\tau_y,~\tau_z)^{\intercal}$ and torques $\vec{\tau}_{\vec{\Theta}}=(\tau_\phi,~\tau_\theta,~\tau_\psi)^{\intercal}$ relative to the body-fixed coordinate frame.

The transformation of body-fixed frame variables to global variables is given by
\begin{align}
    \vec{\dot{p}} = \vec{J}_1(\vec{\Theta}) \vec{v},\\
    \vec{\dot{\Theta}} = \vec{J}_2(\vec{\Theta}) \vec{\omega},
\end{align}
where $\vec{J}_1(\vec{\Theta})$ and $\vec{J}_2(\vec{\Theta})$ are functions of the current Euler angles $\vec{\Theta}$.
The $\vec{\dot{p}}$ denotes the velocity of the \ac{USV} in the global frame, while $\vec{\dot{\Theta}}$ denotes the angular rates of Euler angles.

The transformation $\vec{J}_1(\vec{\Theta})$ defines rotation from a body-fixed frame to global frame as
\begin{align}
    \vec{J}_1(\vec{\Theta}) = \vec{R}_{\psi} \vec{R}_{\theta} \vec{R}_{\phi},
\end{align}
where $\vec{R}_{\psi}$ is rotation around z-axis, $\vec{R}_{\theta}$ is rotation around y-axis, and $\vec{R}_{\phi}$ is rotation around x-axis.
According to \citep{Fossen2011}, the transformation $\vec{J}_2(\vec{\Theta})$ is
\begin{align}
    \vec{J}_2(\vec{\Theta}) = \begin{pmatrix} 
     1 & \quad\sin\phi\tan\theta & \quad\cos\phi\tan\theta\\
     0 & \cos\phi & -\sin\phi\\
     0 & \dfrac{\sin\phi}{\cos\theta} & \dfrac{\cos\phi}{\cos\theta}
    \end{pmatrix}.
\end{align}
However, the transformation matrix $\vec{J}_2(\vec{\Theta})$ is not defined for $\theta=\dfrac{\pi}{2}+k\pi,~k\in\mathbb{Z}$.
A \ac{USV} moving over water will likely not experience such a situation, but nevertheless, the solution to such a problem is to use two conventions of the Euler angles representations with different singularities.
If the \ac{USV} is close to the singularity point in one convention, it is switched to another convention.

Finally, the kinematic transformation is mathematically described as
\begin{align}
    \begin{pmatrix} 
    \vec{\dot{p}}\\ \vec{\dot{\Theta}}
    \end{pmatrix} =
    \begin{pmatrix}
     \vec{J}_1(\vec{\Theta}) & \vec{O}_{3\times3}\\
     \vec{O}_{3\times3} & \vec{J}_2(\vec{\Theta})
    \end{pmatrix}
    \begin{pmatrix}
     \vec{v}\\ \vec{\omega}
    \end{pmatrix},
\end{align}
or, equivalently,
\begin{align}
    \vec{\dot{\eta}} = \vec{J}(\vec{\eta}) \vec{\nu}.
\end{align}

\subsection{Nonlinear USV model} \label{sec:usv_nonlinear_model}
The novel nonlinear 6 \ac{DOF} \ac{USV} model is based on the marine vessel dynamics analysis presented in \citep{Fossen2011, Fossen2002, Fossen1994}.
We adopted the equations and made assumptions detailed below to create a mathematical \ac{USV} model suitable for state estimation tasks.
Moreover, we introduce a novel mathematical model capturing wave dynamics and incorporate it into the \ac{USV} model (\refsec{sec:waves_model}). 
This results in the novel nonlinear 6 \ac{DOF} \ac{USV} model describing \ac{USV} motion on a rough water surface presented in equations \refeq{eq:nonlin_usv_model_first}--\refeq{eq:nonlin_usv_model_last} in \refsec{sec:waves_model}.

We introduce the basic nonlinear 6 \ac{DOF} \ac{USV} model as
\begin{align}
    \vec{\dot{\eta}} &= \vec{J}(\vec{\eta})\vec{\nu},\\
    \vec{\dot{\nu}} &= \vec{M}^{-1} ( \vec{\tau} - \vec{C}(\vec{\nu})\vec{\nu} - \vec{D}(\vec{\nu})\vec{\nu} - \vec{g}(\vec{\eta}) ),
\end{align}
where
\begin{itemize}
    \item $\vec{M} \in \mathbb{R}^{6\times 6}$ represents the inertia matrix,
    \item $\vec{C}(\vec{\nu}) \in \mathbb{R}^{6\times 6}$ denotes the matrix of Coriolis and centripetal terms,
    \item $\vec{D}(\vec{\nu}) \in \mathbb{R}^{6\times 6}$ is the damping matrix,
    \item $\vec{g}(\vec{\eta}) \in \mathbb{R}^{6}$ represents gravitational force and torques,
    \item $\vec{\tau} \in \mathbb{R}^{6}$ denotes the vector forces acting on the \ac{USV}, e.g., wave forces, wind forces, and control inputs.
\end{itemize}
The inertia $\vec{M}$ is composed of two matrices
\begin{align}
    \vec{M} = \vec{M}_{RB} + \vec{M}_{A}. \label{eq:inertia_M}
\end{align}
The matrix $\vec{M}_{RB}$ is the positive definite rigid-body mass matrix
\begin{align}
    \vec{M}_{RB} = \begin{pmatrix}
     m \vec{I}_{3\times3} & \vec{O}_{3\times3}\\
     \vec{O}_{3\times3}  & \vec{I}_{b}
    \end{pmatrix},
\end{align}
where $m$ is the mass of the \ac{USV} and $\vec{I}_{3\times3}$ is the identity matrix.
The $\vec{I}_{b}$ is the inertia matrix, whose components $I_{x},~I_{y},~I_{z}$ are the moments of inertia about  the corresponding body-fixed frame axes $x_b,~y_b,~z_b$ and $I_{xy}=I_{yx}$, $I_{xz}=I_{zx}$, $I_{yz}=I_{zy}$.
The matrix $\vec{M}_A$ is the virtual hydrodynamic added mass caused by a moving object in a fluid.
The matrix $\vec{M}_A$ is expressed as
\begin{align}
    \vec{M}_A = \begin{pmatrix}
     X_{\dot{u}} & X_{\dot{v}} & X_{\dot{w}} & X_{\dot{p}} & X_{\dot{q}} & X_{\dot{r}}\\ 
     Y_{\dot{u}} & Y_{\dot{v}} & Y_{\dot{w}} & Y_{\dot{p}} & Y_{\dot{q}} & Y_{\dot{r}}\\ 
     Z_{\dot{u}} & Z_{\dot{v}} & Z_{\dot{w}} & Z_{\dot{p}} & Z_{\dot{q}} & Z_{\dot{r}}\\ 
     K_{\dot{u}} & K_{\dot{v}} & K_{\dot{w}} & K_{\dot{p}} & K_{\dot{q}} & K_{\dot{r}}\\ 
     M_{\dot{u}} & M_{\dot{v}} & M_{\dot{w}} & M_{\dot{p}} & M_{\dot{q}} & M_{\dot{r}}\\ 
     N_{\dot{u}} & N_{\dot{v}} & N_{\dot{w}} & N_{\dot{p}} & N_{\dot{q}} & N_{\dot{r}}
    \end{pmatrix},
\end{align}
where elements $X_{\dot{u}},~X_{\dot{v}},\ldots,~N_{\dot{r}}$ are coefficients of hydrodynamic added mass.

The Coriolis and centripetal matrix $\vec{C}(\vec{\nu})$ is similar to the inertia matrix $\vec{M}$ \refeq{eq:inertia_M} composed of two matrices
\begin{align}
    \vec{C}(\vec{\nu}) = \vec{C}_{RB}(\vec{\nu}) + \vec{C}_A(\vec{\nu}).
\end{align}
The Coriolis and the centripetal term is a result of the rotation of the body-fixed reference frame.
The $\vec{C}_{RB}$ stands for rigid-body Coriolis and centripetal matrix.
In order to move the rigid body in a fluid, the hydrodynamic Coriolis and centripetal matrix $\vec{C}_A(\vec{\nu})$ is defined as
\begin{align}
    \vec{C}_A(\vec{\nu}) = 
    \begin{pmatrix}
     0 & 0 & 0 & 0 & -C_3 & C_2\\
     0 & 0 & 0 & C_3 & 0 & -C_1\\
     0 & 0 & 0 & -C_2 & C_1 & 0\\
     0 & -C_3 & C_2 & 0 & -C_6 & C_5\\
     C_3 & 0 & -C_1 & C_6 & 0 & -C_4\\
     -C_2 & C_1 & 0 & -C_5 & C_4 & 0
    \end{pmatrix},
\end{align}
where the individual elements of the matrix $\vec{C}_A(\vec{\nu})$ are defined as
\begin{align}
    C_1 &= X_{\dot{u}}u + X_{\dot{v}}v + X_{\dot{w}}w + X_{\dot{p}}p + X_{\dot{q}}q + X_{\dot{r}}r,\\ 
    C_2 &= Y_{\dot{u}}u + Y_{\dot{v}}v + Y_{\dot{w}}w + Y_{\dot{p}}p + Y_{\dot{q}}q + Y_{\dot{r}}r,\\ 
    C_3 &= Z_{\dot{u}}u + Z_{\dot{v}}v + Z_{\dot{w}}w + Z_{\dot{p}}p + Z_{\dot{q}}q + Z_{\dot{r}}r,\\ 
    C_4 &= K_{\dot{u}}u + K_{\dot{v}}v + K_{\dot{w}}w + K_{\dot{p}}p + K_{\dot{q}}q + K_{\dot{r}}r,\\ 
    C_5 &= M_{\dot{u}}u + M_{\dot{v}}v + M_{\dot{w}}w + M_{\dot{p}}p + M_{\dot{q}}q + M_{\dot{r}}r,\\ 
    C_6 &= N_{\dot{u}}u + N_{\dot{v}}v + N_{\dot{w}}w + N_{\dot{p}}p + N_{\dot{q}}q + N_{\dot{r}}r.
\end{align}

The term $\vec{D}(\vec{\nu})$ represents the damping of the system that is expressed as linear damping 
\begin{align}
    \vec{D}(\vec{\nu}) \approx \vec{D} = \begin{pmatrix}
     X_u & 0 & 0 & 0 & 0 & 0\\
     0 & Y_v & 0 & Y_p & 0 & Y_r\\
     0 & 0 & Z_w & 0 & Z_q & 0\\
     0 & K_v & 0 & K_p & 0 & K_r\\
     0 & 0 & M_w & 0 & M_q & 0\\
     0 & N_v & 0 & N_p & 0 & N_r\\
    \end{pmatrix}, \label{eq:nonlinear_damping_term}
\end{align}
where the coefficients $X_u$, $Y_v,\ldots,~N_r$ are known as hydrodynamic derivatives.
There are several causes of the damping --- potential damping, surface friction, wave drift damping, damping due to vortex shedding, and lifting forces.

It is assumed that gravitational forces and torques $\vec{g}(\vec{\eta})$, also called restoring forces, are zero in equilibrium position $z=0$, in which nominal water volume is displaced by the \ac{USV}.
The function $\vec{g}(\vec{\eta})$ is rewritten using linear approximation as
\begin{align}
    \vec{g}(\vec{\eta}) \approx \vec{G}\vec{\eta}, \label{eq:restoring_forces_def}
\end{align}
where matrix $\vec{G}$ has the following form
\begin{align}
    \vec{G} = \begin{pmatrix}
     0 & 0 & 0 & 0 & 0 & 0\\
     0 & 0 & 0 & 0 & 0 & 0\\
     0 & 0 & -Z_{z} & 0 & -Z_{\theta} & 0\\
     0 & 0 & 0 & -K_{\phi} & 0 & 0\\
     0 & 0 & -M_{z} & 0 & -M_{\theta} & 0\\
     0 & 0 & 0 & 0 & 0 & 0\\
    \end{pmatrix},
\end{align}
and $Z_{z}$, $Z_{\theta}$, $K_{\phi}$, $M_{z}$, and $M_{\theta}$ are the gravitational coefficients.

The modeling and measurement of the vector of forces $\vec{\tau}$ acting on the \ac{USV} are challenging.
The effects of waves are presented in \refsec{sec:waves_model}, as it is one of the main challenges for the state estimation approach designed in this paper.
Assuming that there are no acting forces $\vec{\tau}$, the nonlinear model of the \ac{USV} is expressed as
\begin{align}
    \vec{\dot{\eta}} &= \vec{J}(\vec{\eta})\vec{\nu},\label{eq:general_nonlinear_usv_model_eta}\\
    \vec{\dot{\nu}} &= \vec{M}^{-1} \left( -\vec{C}(\vec{\nu})\vec{\nu} -\vec{D}\vec{\nu} -\vec{G}\vec{\eta} \right).\label{eq:general_nonlinear_usv_model_nu}
\end{align}

\subsection{Linear USV model} \label{sec:usv_linear_model}
The nonlinear \ac{USV} state space model defined in \refeq{eq:general_nonlinear_usv_model_eta} and \refeq{eq:general_nonlinear_usv_model_nu} is linearized under the following assumptions.
Firstly, the roll angle $\phi$ and pitch angle $\theta$ are assumed to be small.
This assumption also holds for a \ac{USV} whose roll and pitch motions are limited.
The \refeq{eq:general_nonlinear_usv_model_eta} is reformulated using the first assumption as
\begin{align}
    \vec{\dot{\eta}} = \vec{J}(\vec{\eta})\vec{\nu} \overset{\phi=\theta=0}{\approx} \vec{J}_{\psi}(\psi)\vec{\nu}, \label{eq:usv_linear_model_1_assumption} 
\end{align}
where transformation matrix $\vec{J}_{\psi}(\psi)$ is
\begin{align}
    \vec{J}_{\psi}(\psi) = \begin{pmatrix}
     \vec{R}_{\psi} & \vec{O}_{3\times3}\\
     \vec{O}_{3\times3} & \vec{I}_{3\times3}
    \end{pmatrix}.
\end{align}

Using \refeq{eq:usv_linear_model_1_assumption}, the \textit{Vessel parallel coordinate system} is defined as
\begin{align}
    \vec{\eta}_L = \vec{J}_{\psi}^{\intercal}(\psi) \vec{\eta}, \label{eq:vessel_parallel_coords}
\end{align}
where $\vec{\eta}_L$ denotes the position and orientation in the global coordinate frame expressed in the body-fixed coordinate frame and $\vec{J}_{\psi}^{\intercal}(\psi)\vec{J}_{\psi}(\psi)=\vec{I}_{6\times6}$.
The time derivative of $\vec{\eta}_L$ is expressed as
\begin{align}
    \vec{\dot{\eta}}_L &= \vec{\dot{J}}_{\psi}^{\intercal}(\psi) \vec{\eta} + \vec{J}_{\psi}^{\intercal}(\psi) \vec{\dot{\eta}}. \label{eq:eta_L_time_derivative}
\end{align}
After substitution of the term $\vec{\eta}=\vec{J}_{\psi}(\psi) \vec{\eta}_L$ and $\vec{\dot{\eta}}\approx \vec{J}_{\psi}(\psi)\vec{\nu}$, the equation \refeq{eq:eta_L_time_derivative} becomes
\begin{align}
    \vec{\dot{\eta}}_L &= \vec{\dot{J}}_{\psi}^{\intercal}(\psi) \vec{J}_{\psi}(\psi) \vec{\eta}_L + \vec{J}_{\psi}^{\intercal}(\psi) \vec{J}_{\psi}(\psi)\vec{\nu} = r\vec{S}\vec{\eta}_L + \vec{\nu} \label{eq:n_vessel_coords_nonlinear},
\end{align}
where $r$ is the yaw angular velocity and
\begin{align}
    \vec{S} = \begin{pmatrix}
     0 & 1 & 0 & 0 & 0 & 0\\ 
     -1 & 0 & 0 & 0 & 0 & 0\\ 
     0 & 0 & 0 & 0 & 0 & 0\\
     0 & 0 & 0 & 0 & 0 & 0\\
     0 & 0 & 0 & 0 & 0 & 0\\
     0 & 0 & 0 & 0 & 0 & 0\\
    \end{pmatrix}.
\end{align}
Considering $r \approx 0$, the equation \refeq{eq:n_vessel_coords_nonlinear} is as 
\begin{align}
    \vec{\dot{\eta}}_L \approx \vec{\nu}.
\end{align}

The term $\vec{G}\vec{\eta}$ \refeq{eq:restoring_forces_def} representing the gravitational and buoyancy forces is also expressed using the Vessel parallel coordinate system as 
\begin{align}
    \vec{G}\vec{\eta} \overset{\phi=\theta=0}{\approx} \vec{G}\vec{\eta}_L.
\end{align}
The nonlinear damping term $\vec{D}(\vec{\nu})$ is converted to a linear form as presented in \refeq{eq:nonlinear_damping_term}.
The last nonlinear term represents Coriolis and centripetal forces $\vec{C}(\vec{\nu})$.
Assuming that $\phi=\theta=0$ and that, in low-speed applications, $\vec{\nu}\approx\vec{0}$, the term $\vec{C}(\vec{\nu})$ becomes the zero matrix $\vec{C}(\vec{\nu})=\vec{O}_{6\times6}$.
Finally, the nonlinear state space equations \refeq{eq:general_nonlinear_usv_model_eta} and \refeq{eq:general_nonlinear_usv_model_nu} are transformed into a linear form as
\begin{align}
   \vec{\dot{\eta}}_L &= \vec{\nu},\\
    \vec{\dot{\nu}} &= -\vec{M}^{-1}\vec{D}\vec{\nu} - \vec{M}^{-1}\vec{G}\vec{\eta}_L.
\end{align}
The \ac{LTI} state space model is expressed as
\begin{align}
    \vec{\dot{x}}_{vp} = \vec{A}_{vp}\vec{x}_{vp},\label{eq:linear_model_usv}
\end{align}
where $\vec{x}_{vp} = (\vec{\eta}_{L}^{\intercal},~\vec{\nu}^{\intercal})^{\intercal}$ and 
\begin{align}
    \vec{A}_{vp} = \begin{pmatrix}
     \vec{O}_{6\times6} & \vec{I}_{6\times6}\\
     -\vec{M}^{-1}\vec{G} & -\vec{M}^{-1}\vec{D}
    \end{pmatrix}.
\end{align}
The global position $\vec{\eta}$ is computed from $\vec{\eta}_L$ as
\begin{align}
    \vec{\eta} = \vec{J}_{\psi}(\psi) \vec{\eta}_L.
\end{align}

\section{WAVE DYNAMICS IN USV MODEL}\label{sec:waves_model}
The motion of the \ac{USV} is significantly influenced by the wave forces.
The results of their actions are oscillatory motions with a zero mean \citep{Fossen2011}.
Under frequency decomposition, the wave elevation at time $t$ is expressed as a sum of $N\in\mathbb{Z}^{+}$ harmonic components
\begin{align}
    \zeta(t) = \sum_{k=1}^{N} A_k \cos(\omega_k t + \epsilon_k),
\end{align}
where $A_k$ is the amplitude of the wave component $k$, $\omega_k$ denotes the frequency of wave component $k$, and $\epsilon_k$ is a random phase angle of wave component $k$.
The characteristics of waves are captured in their spectra $S(\omega_k)$~\citep{Fossen2011}.
The amplitude $A_k$ of wave component $k$ is related to the wave spectrum $S(\omega_k)$ as
\begin{align}
    \dfrac{1}{2}A_k^2 = S(\omega_k)\Delta \omega_k,
\end{align}
where $\Delta \omega_k$ denotes a constant difference between the frequencies of wave components $k$ and $k-1$.

\subsection{Nonlinear USV model with wave dynamics}
\label{sec:waves_model_nonlin}
To use wave elevation in the nonlinear \ac{USV} model, a system generating the harmonic signal as one wave component is needed.
We propose to design the system of one wave component as
\begin{align}
    \dot{x}_{\omega_{N1}} &= x_{\omega_{N2}},\label{eq:nonlinear_wave_component_eq_first}\\
    \dot{x}_{\omega_{N2}} &= -x_{\omega_{N3}}\sin(x_{\omega_{N1}}) -\gamma x_{\omega_{N2}},\\
    \dot{x}_{\omega_{N3}} &= 0,\\
    y_{\omega_{N}} &= x_{\omega_{N2}},\label{eq:nonlinear_wave_component_eq_last}
\end{align}
where $x_{\omega_{N1}},~x_{\omega_{N2}}$, and $x_{\omega_{N3}}$ are state variables of the system, $y_{\omega_{N}}$ is an output signal of one wave component, and $\gamma$ is a damping term of the wave component.
The state $x_{\omega_{N3}}$ corresponds to the frequency of the wave component that does not evolve in time.
The system defined by equations \refeq{eq:nonlinear_wave_component_eq_first}--\refeq{eq:nonlinear_wave_component_eq_last} is expressed using $\vec{x}_{\omega_{N}} = (x_{\omega_{N1}},~x_{\omega_{N2}},~x_{\omega_{N3}})^{\intercal}$ as
\begin{align}
    \vec{\dot{x}}_{\omega_{N}} &= \vec{f}_{\omega_{N}}(\vec{x}_{\omega_{N}}),\\
    y_{\omega_{N}} &= \vec{g}_{\omega_{N}}(\vec{x}_{\omega_{N}}).\label{eq:nonlinear_wave_component_system}
\end{align}
The $N_{nc}\in\mathbb{Z}^{+}$ individual components $y_{\omega_{N}}$ defined in \refeq{eq:nonlinear_wave_component_system} are summed together to obtain the complex wave motion
\begin{align}
    \vec{\dot{x}}_{\omega_{N1}} &= \vec{f}_{\omega_{N}}(\vec{x}_{\omega_{N1}}),\\
    y_{\omega_{N1}} &= \vec{g}_{\omega_{N}}(\vec{x}_{\omega_{N1}}),\\
    &~\vdots\nonumber\\
    \vec{\dot{x}}_{\omega_{NN_{nc}}} &= \vec{f}_{\omega_{N}}(\vec{x}_{\omega_{NN_{nc}}}),\\
    y_{\omega_{NN_{nc}}} &= \vec{g}_{\omega_{N}}(\vec{x}_{\omega_{NN_{nc}}}),\\
    y_{wave} &= y_{\omega_{N1}} + \ldots + y_{\omega_{NN_{nc}}},
\end{align}
which is simplified using a new state $\vec{x}_{wave}$
\begin{align}
    \vec{x}_{wave} &=(\vec{x}^{\intercal}_{\omega_{N1}},\ldots,~\vec{x}^{\intercal}_{\omega_{NN_{nc}}})^{\intercal},\\
    \vec{\dot{x}}_{wave} &= \vec{f}_{wave}(\vec{x}_{wave}),\label{eq:one_wave_system_x}\\
    y_{wave} &= \vec{g}_{wave}(\vec{x}_{wave}).\label{eq:one_wave_system_y}
\end{align}
The $y_{wave}$ \refeq{eq:one_wave_system_y} is used to create a new state vector $\vec{\nu}_{wave}$ as
\begin{align}
    \vec{\nu}_{wave} = (y_{wave,u},~y_{wave,v},~y_{wave,w},\nonumber\\y_{wave,p},~y_{wave,q},~y_{wave,r})^{\intercal},
\end{align}
where each element $y_{wave,u}$, $y_{wave,v}$, $y_{wave,w}$, $y_{wave,p}$, $y_{wave,q}$, $y_{wave,r}$ corresponds to a one wave system defined in \refeq{eq:one_wave_system_x} and \refeq{eq:one_wave_system_y}.
Finally, we propose the novel nonlinear 6 \ac{DOF} model of the \ac{USV} containing wave dynamics as
\begin{align}
    \vec{\dot{\eta}} &= \vec{J}(\vec{\eta})\vec{\nu},\label{eq:nonlin_usv_model_first}\\
    \vec{\dot{\nu}} &= \vec{M}^{-1} \left( -\vec{C}(\vec{\nu})\vec{\nu} -\vec{D}(\vec{\nu})\vec{\nu} -\vec{G}\vec{\eta} \right) + \vec{\nu}_{wave},\\
    \vec{\dot{x}}_{wave,u} &= \vec{f}_{wave}(\vec{x}_{wave,u}),\\
    \vec{\dot{x}}_{wave,v} &= \vec{f}_{wave}(\vec{x}_{wave,v}),\\
    \vec{\dot{x}}_{wave,w} &= \vec{f}_{wave}(\vec{x}_{wave,w}),\\
    \vec{\dot{x}}_{wave,p} &= \vec{f}_{wave}(\vec{x}_{wave,p}),\\
    \vec{\dot{x}}_{wave,q} &= \vec{f}_{wave}(\vec{x}_{wave,q}),\\
    \vec{\dot{x}}_{wave,r} &= \vec{f}_{wave}(\vec{x}_{wave,r}).\label{eq:nonlin_usv_model_last}
\end{align}
Let us define 
\begin{align}
    \vec{x}_{USV} = (\vec{\eta}^{\intercal},~\vec{\nu}^{\intercal},~\vec{x}_{wave,u}^{\intercal},\ldots,~\vec{x}_{wave,r}^{\intercal})^{\intercal}.
\end{align}
Then the nonlinear model of the \ac{USV} defined in \refeq{eq:nonlin_usv_model_first}--\refeq{eq:nonlin_usv_model_last} is written as
\begin{align}
    \vec{\dot{x}}_{USV} = \vec{f}_{USV}(\vec{x}_{USV}). \label{eq:general_nonlinear_usv_model}
\end{align}

\subsection{Linear USV model with wave dynamics}
\label{sec:waves_model_lin}
The wave system defined in \refeq{eq:one_wave_system_x} and \refeq{eq:one_wave_system_y} cannot be used for the linear \ac{USV} model as the wave system is nonlinear.
Therefore, we present the linear state-space model of a one wave component including two states $x_{\omega_{L1}},~x_{\omega_{L2}}$ as follows:
\begin{align}
    \begin{pmatrix}
     \dot{x}_{\omega_{L1}}\\ \dot{x}_{\omega_{L2}}
     \end{pmatrix} =& 
     \begin{pmatrix}
      0 & 1\\
      -\omega_{0_L}^2 & -2\lambda_{L}\omega_{0_L}
    \end{pmatrix}
    \begin{pmatrix}
    x_{\omega_{L1}}\\ x_{\omega_{L2}}
    \end{pmatrix}, \label{eq:linear_wave_model_x}
    \\
    y_{\omega_{L}} &= \begin{pmatrix}
    0 & 1
    \end{pmatrix}
    \begin{pmatrix}
    x_{\omega_{L1}}\\ x_{\omega_{L2}}
    \end{pmatrix},
    \label{eq:linear_wave_model_y}
\end{align}
where $\omega_{0_L}$ represents the frequency of the wave component and $\lambda_{L}$ is the damping of the wave component.
The wave component defined in \refeq{eq:linear_wave_model_x} and \refeq{eq:linear_wave_model_y} is expressed in matrix form as
\begin{align}
    \vec{\dot{x}}_{\omega_L} &= \vec{A}_{\omega_L}\vec{x}_{\omega_L},\\
    y_{\omega_L} &= \vec{C}_{\omega_L}\vec{x}_{\omega_L}.
\end{align}
To achieve complex wave motion composed of several harmonics, the $N_{lc}\in\mathbb{Z}^{+}$ state-space wave components $y_{\omega_L}$ can be joined together with different parameters $\omega_{0_L}$ and $\lambda_{L}$ as
\begin{align}
    \vec{\dot{x}}_{\omega_{L1}} &= \vec{A}_{\omega_{L1}}\vec{x}_{\omega_{L1}},\label{eq:linear_wave_system_first}\\
    y_{\omega_{L1}} &= \vec{C}_{\omega_{L1}}\vec{x}_{\omega_{L1}},\\
    &~\vdots\nonumber\\
    \vec{\dot{x}}_{\omega_{LN_{lc}}} &= \vec{A}_{\omega_{LN_{lc}}}\vec{x}_{\omega_{LN_{lc}}},\\
    y_{\omega_{LN_{lc}}} &= \vec{C}_{\omega_{LN_{lc}}}\vec{x}_{\omega_{LN_{lc}}},\\
    y_{\text{wave}_L} &= y_{\omega_{L1}} + \ldots + y_{\omega_{LN_{lc}}}.\label{eq:linear_wave_system_last}
\end{align}
The system described in \refeq{eq:linear_wave_system_first}--\refeq{eq:linear_wave_system_last} is expressed as one linear system
\begin{align}
    \vec{\dot{x}}_{wave_{L}} &= \vec{A}_{wave_{L}} \vec{x}_{wave_{L}},\label{eq:linear_wave_system_x}\\
    y_{\text{wave}_L} &= \vec{C}_{wave_{L}} \vec{x}_{wave_{L}},\label{eq:linear_wave_system_y}
\end{align}
where 
\vspace{-5pt}
\begin{align}
    \vec{x}_{wave_{L}} &= (\vec{x}_{\omega_{L1}}^{\intercal},\ldots,~\vec{x}_{\omega_{LN_{lc}}}^{\intercal} )^{\intercal},\\
    \vec{A}_{wave_{L}} &= \diag\{ \vec{A}_{\omega_{L1}},\ldots,~ \vec{A}_{\omega_{LN_{lc}}}\},\label{eq:A_wave_L}\\
    \vec{C}_{wave_{L}} &= \begin{pmatrix}
    \vec{C}_{\omega_{L1}} & \cdots & \vec{C}_{\omega_{LN_{lc}}}
    \end{pmatrix}.
\end{align}
In \refeq{eq:A_wave_L} $\diag\{\cdot\}$ represents a block diagonal matrix of the given elements $\vec{A}_{\omega_{L1}},\ldots, ~\vec{A}_{\omega_{LN_{lc}}}$.

The one wave system in \refeq{eq:linear_wave_system_x} and \refeq{eq:linear_wave_system_y} influences one \ac{USV} state of $\vec{\nu}$ \refeq{eq:usv_body_states}.
Therefore, we propose the complex wave system for the \ac{USV} states $\vec{\nu}$ as
\begin{align}
    \vec{\dot{x}}_{wave,\vec{\nu}} &= \vec{A}_{wave,\vec{\nu}}\vec{x}_{wave,\vec{\nu}},\label{eq:complex_wave_system_1}\\
    \vec{y}_{wave,\vec{\nu}} &= \vec{C}_{wave,\vec{\nu}}\vec{x}_{wave,\vec{\nu}},\\
    \vec{A}_{wave,\vec{\nu}} &= \diag\{ \vec{A}_{wave_{L}},~\vec{A}_{wave_{L}},~\vec{A}_{wave_{L}},\nonumber\\&\vec{A}_{wave_{L}},~\vec{A}_{wave_{L}},~\vec{A}_{wave_{L}} \}, \\
    \vec{C}_{wave,\vec{\nu}} &= 
      \left( \begin{matrix}
        \vec{C}_{wave_{L}} & \vec{O}_{1\times2N_{lc}} & \vec{O}_{1\times2N_{lc}} \\ 
         \vec{O}_{1\times2N_{lc}} & \vec{C}_{wave_{L}} & \vec{O}_{1\times2N_{lc}} \\ 
         \vec{O}_{1\times2N_{lc}} & \vec{O}_{1\times2N_{lc}} & \vec{C}_{wave_{L}} \\ 
         \vec{O}_{1\times2N_{lc}} & \vec{O}_{1\times2N_{lc}} & \vec{O}_{1\times2N_{lc}} \\ 
         \vec{O}_{1\times2N_{lc}} & \vec{O}_{1\times2N_{lc}} & \vec{O}_{1\times2N_{lc}} \\ 
         \vec{O}_{1\times2N_{lc}} & \vec{O}_{1\times2N_{lc}} & \vec{O}_{1\times2N_{lc}} \\ 
    \end{matrix}
    \right.\nonumber
    \\
    &\hfill\left. \begin{matrix}
        \vec{O}_{1\times2N_{lc}} & \vec{O}_{1\times2N_{lc}}& \vec{O}_{1\times2N_{lc}}\\
        \vec{O}_{1\times2N_{lc}} & \vec{O}_{1\times2N_{lc}}& \vec{O}_{1\times2N_{lc}}\\
        \vec{O}_{1\times2N_{lc}} & \vec{O}_{1\times2N_{lc}}& \vec{O}_{1\times2N_{lc}}\\
        \vec{C}_{wave_{L}} & \vec{O}_{1\times2N_{lc}}& \vec{O}_{1\times2N_{lc}}\\
        \vec{O}_{1\times2N_{lc}} &\vec{C}_{wave_{L}} & \vec{O}_{1\times2N_{lc}}\\
        \vec{O}_{1\times2N_{lc}} & \vec{O}_{1\times2N_{lc}}& \vec{C}_{wave_{L}}\\
    \end{matrix}
    \right)\label{eq:complex_wave_system_2},
\end{align}
where $\vec{A}_{wave,\vec{\nu}}$ is a block diagonal matrix of this complex wave system and 
\begin{align}
    \vec{x}_{wave,\vec{\nu}} = (\vec{x}_{wave_{L},u}^{\intercal},~\vec{x}_{wave_{L},v}^{\intercal},~\vec{x}_{wave_{L},w}^{\intercal}, \nonumber\\\vec{x}_{wave_{L},p}^{\intercal},~\vec{x}_{wave_{L},q}^{\intercal},~\vec{x}_{wave_{L},r}^{\intercal})^{\intercal}.
\end{align}
We present a linear 6 \ac{DOF} model of the \ac{USV}  containing wave dynamics \refeq{eq:complex_wave_system_1}-\refeq{eq:complex_wave_system_2} as
\begin{align}
    \vec{\dot{x}}_{L,usv} = \vec{A}_{L,usv}\vec{x}_{L,usv},\label{eq:linear_model_usv_wave}
\end{align}
where $\vec{x}_{L,usv} = (\vec{\eta}_{L}^{\intercal},~\vec{\nu}^{\intercal},~\vec{x}^{\intercal}_{wave,\vec{\nu}})^{\intercal}$ and 
\begin{align}
    \vec{A}_{L,usv} = \begin{pmatrix}
     \vec{O}_{6\times6} & \vec{I}_{6\times6} & \vec{O}_{6\times12N_{lc}}\\
     -\vec{M}^{-1}\vec{G} & -\vec{M}^{-1}\vec{D} & \vec{C}_{wave,\vec{\nu}}\\
     \vec{O}_{12N_{lc}\times6} & \vec{O}_{12N_{lc}\times6} & \vec{A}_{wave,\vec{\nu}}
    \end{pmatrix}.
\end{align}

\subsection{State estimator}
\label{sec:kalman_filters}
The \ac{UKF} \citep{Julier1997728} with the novel nonlinear \ac{USV} model \refeq{eq:general_nonlinear_usv_model} was selected as a state estimator.
The main advantage of this filter is that no linearization step is required.
The filter uses a nonlinear model of system dynamics, as well as nonlinear models of sensors, and consists of two steps:
\begin{itemize}
    \item \textit{prediction step} --- propagate the current state and its covariance through a model of the system,
    \item \textit{correction step} --- use obtained measurement to update the current state and its covariance.
\end{itemize}

The parameters of the Kalman filter are the covariance matrices for the system and sensors, corresponding to the process noise of the model and the measurement noise from the sensors.
The covariance matrix of the process noise is estimated using simulation data by measuring deviations between predicted states and ground-truth states over time and computing their covariance.
The sensors' covariance matrices are estimated by collecting data from the sensors and calculating the variance of repeated measurements.
Finally, covariance matrices need to be fine-tuned to achieve the desired performance, noise reduction, and delay suitable for our \ac{UAV} controller during landing.
To simplify the fine-tuning process, we kept all covariance matrices diagonal.

The proposed novel estimation approach repeatedly applies the prediction step of the \ac{UKF} to compute the desired number of predictions representing the future \ac{USV} states.
Firstly, the novel nonlinear mathematical \ac{USV} model $\vec{f}_{usv}$ defined in \refeq{eq:general_nonlinear_usv_model} is discretized
using Runge-Kutta method of the fourth order
\begin{align}
     \vec{x}_{usv}(k_t+1) &= \vec{f}_{usv,d}(\vec{x}_{usv}(k_t), k_t),
\end{align}
where $k_t$ is a discrete time step.
The first predicted state $\vec{\hat{x}}_{usv}(0)$ and its covariance matrix $\vec{\hat{P}}_{usv}(0)$ are initialized by the latest estimated state $\vec{x}_{usv}(t_{cur})$ in time $t_{cur}$ and its covariance matrix $\vec{P}_{usv}(t_{cur})$ in $t_{cur}$
\begin{align}
    \vec{\hat{x}}_{usv}(0) = \vec{x}_{usv}(t_{cur}),\\
    \vec{\hat{P}}_{usv}(0) = \vec{P}_{usv}(t_{cur}).
\end{align}
Finally, the desired number of \ac{USV} state predictions $N_{p,N}\in\mathbb{Z}^+$ is computed using the prediction step of \ac{UKF} $\vec{g}_{ukf}$ as
\begin{align} 
    (\vec{\hat{x}}_{usv}(k_{p,N}+1),~\vec{\hat{P}}_{usv}(k_{p,N}+1)) =\nonumber\\\vec{g}_{ukf}(\vec{\hat{x}}_{usv}(k_{p,N}),~\vec{\hat{P}}_{usv}(k_{p,N})),
\end{align}
where $k_{p,N}=0,1,\ldots,N_{p,N}-1$.
The approach using \ac{UKF} and the proposed novel nonlinear \ac{USV} model extended by the proposed nonlinear wave model \refeq{eq:general_nonlinear_usv_model} is denoted as \usvesnonlinear{} in the experimental part of this paper.
For comparison, we also present a linear version of the proposed approach, which uses \ac{LKF} \citep{kalman1960new} with the discrete form $\vec{\bar{A}}_{L,usv}$ of the linear \ac{USV} model extended by a linear wave model \refeq{eq:linear_model_usv_wave}, denoted as \usveslinear{}.
Similarly to the \usvesnonlinear{}, the prediction step of the \ac{LKF} is repeatedly applied to predict desired number $N_{p,L}\in\mathbb{Z}^+$ of future \ac{USV} states $\vec{\hat{x}}_{L,usv}(k_{p,L})$ and corresponding covariance matrices $\vec{\hat{P}}_{L,usv}(k_{p,L})$
\begin{align}
    \vec{\hat{x}}_{L,usv}(k_{p,L}+1) &= \vec{\bar{A}}_{L,usv} \vec{\hat{x}}_{L,usv}(k_{p,L}),\\
    \vec{\hat{P}}_{L,usv}(k_{p,L}+1) &= \vec{\bar{A}}_{L,usv} \vec{\hat{P}}_{L,usv}(k_{p,L}) \vec{\bar{A}}_{L,usv}^{\intercal} + \vec{Q}_{L,usv},
\end{align}
where $\vec{Q}_{L,usv}\in\mathbb{R}^{12(1+N_{lc})\times12(1+N_{lc})}$ is system noise matrix, $k_{p,L}=0,1,\ldots,N_{p,L}-1$, and the first predicted state $\vec{\hat{x}}_{L,usv}(0)$ and its covariance matrix $\vec{\hat{P}}_{L,usv}(0)$ are initialized by the latest estimated state $\vec{x}_{L,usv}(t_{cur})$ in time $t_{cur}$ and its covariance matrix $\vec{P}_{L,usv}(t_{cur})$ in $t_{cur}$
\begin{align}
    \vec{\hat{x}}_{L,usv}(0)&=\vec{x}_{L,usv}(t_{cur}),\\
    \vec{\hat{P}}_{L,usv}(0)&=\vec{P}_{L,usv}(t_{cur}).
\end{align}

\section{SENSORS FOR EXPERIMENTAL EVALUATION}
\label{sec:sensors}
The Kalman filter selected as a state estimator (\refsec{sec:kalman_filters}) requires sensor data obtained in real-time to estimate \ac{USV} states.
The sensors are divided into two groups: 1) sensors that are directly placed on the \ac{USV}, 2) \ac{UAV} onboard sensors.
All sensor data are fused together to obtain an accurate estimation of \ac{USV} states.

\subsection{Sensors placed on USV} \label{sec:sensors_placed_on_USV}
In many real-world scenarios, the \ac{UAV} is at such a distance that its onboard sensors do not yield usable data for \ac{USV} state estimation.
However, it is assumed that the communication link between the \ac{UAV} and the \ac{USV} is present.
To move the \ac{UAV} towards the \ac{USV}, the position of the \ac{USV} must at least be estimated.
The \ac{GNSS} receiver is a widely used device to estimate the global position \citep{gps_nav, gps_landing_antennas, RobustGPSloc, WENDEL2006527, gps_time}.

The global position obtained from the \ac{GNSS} receiver placed on the \ac{USV} is sent to the \ac{UAV} via a wireless communication link.
The \ac{UAV} uses this received data to roughly estimate the position of the \ac{USV}. 
The control system takes the \ac{UAV} to the proximity of the \ac{USV}, where onboard sensors are used to update the estimates to be more accurate (\refsec{sec:UAV_onboard_sensors}).
The \ac{UAV} uses the received \ac{GNSS} data from the \ac{USV} in the correction step of the Kalman filters (\refsec{sec:kalman_filters})

\begin{figure}[!tb]
  \centering
  \includegraphics[width=\linewidth]{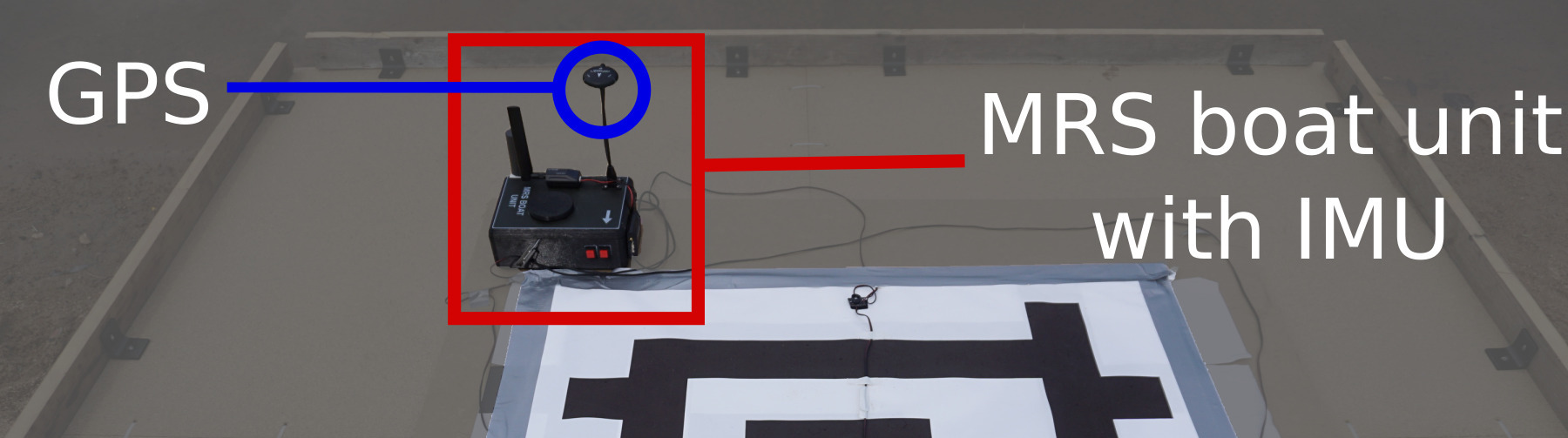}
  \caption{The GPS and the IMU placed onboard the \ac{USV}.}
  \label{fig:landing_board_tmp}
\end{figure}

Another sensor on the \ac{USV} is the \ac{IMU}.
The \ac{IMU} measures heading, angular velocity and linear acceleration.
The \ac{IMU} and the \ac{GPS} are part of a purposely design \ac{MRS} boat unit within our group\footnote{\url{https://mrs.fel.cvut.cz/}\label{footnote:mrs_webpage}} containing the necessary electronic equipment for the \ac{USV} board.
A photo of the \ac{USV} board is shown in \reffig{fig:landing_board_tmp}.
Similar to the \ac{GPS} data, the \ac{IMU} data are sent to the \ac{UAV} via a wireless communication link.
The received \ac{IMU} data are used in the correction step of the Kalman filters (\refsec{sec:kalman_filters}).
The motivation for adding the \ac{IMU} sensor on the \ac{USV} board is to obtain a better estimation accuracy for the motion caused by the waves (\refsec{sec:waves_model}).
Moreover, the combination of \ac{IMU} and \ac{GPS} results in a more accurate estimate of the system states --- position, orientation, velocity, and angular velocity.

\subsection{UAV onboard sensors} \label{sec:UAV_onboard_sensors}
The \refsec{sec:sensors_placed_on_USV} presents the sensors placed on the \ac{USV}, whose data have to be sent to the \ac{UAV} via a communication link.
However, reliable data exchange with minimal latency between the \ac{USV} and the \ac{UAV} is challenging to achieve in real-world environments \citep{TRAN201967}.
In order to cope with the issues of not perfect communication (\refsec{sec:sensors_placed_on_USV}) or \ac{USV} sensors faults, the \ac{UAV} is equipped with onboard vision systems that allow for direct estimation of the \ac{USV} states without the need for communication or a common reference frame between the \ac{USV} and the \ac{UAV}.
To increase the redundancy and ensure a properly working system in varying real-world conditions, two onboard sensors are used: an \ac{UVDAR} system \citep{walter_icra2020, uvdar_dirfol, uvdd2, uvdd1} and an AprilTag detector \citep{krogius2019iros, wang2016iros, olson2011tags}.
Both vision-based systems require dedicated markers placed on the target.

The AprilTags detected by the AprilTag detector (\refsec{sec:AprilTag_sensor}) are passive reflective markers. 
Therefore, they are only useful under lighting conditions sufficient for imaging.
If the \ac{UAV} has to land in the dark, the AprilTag detector will not produce a reliable position and orientation estimation.
However, the \ac{UVDAR} system (\refsec{sec:UVDAR_sensor}) is able to work properly at any time of day due to the active blinking markers.
Since achieving robustness is important in intended missions, we demonstrated the possibility of using two complementary onboard detection systems.
Both systems are described in the \refsec{sec:UVDAR_sensor} and \ref{sec:AprilTag_sensor}.

\subsection{UVDAR system} 
\label{sec:UVDAR_sensor}
The \ac{UVDAR} system is an onboard vision-based relative localization system developed by the \ac{MRS} group\textsuperscript{\ref{footnote:mrs_webpage}} at \ac{CTU} in Prague.
One of the great advantages of this system is that no communication is needed to obtain estimations of states of objects in the \ac{UAV} proximity.
As mentioned above, the \ac{UVDAR} system requires markers placed on the target.
The~marker is composed of a \ac{UV} LED that blinks a unique binary signal code.
The~blinking \ac{UV} LEDs are captured by a \ac{UV}-only-sensitive camera placed onboard the \ac{UAV}.
The principle of the \ac{UVDAR} system is presented in detail in \citep{walter_icra2020, uvdar_dirfol, uvdd2, uvdd1}.

\begin{figure}[!tb]
  \centering
  \input{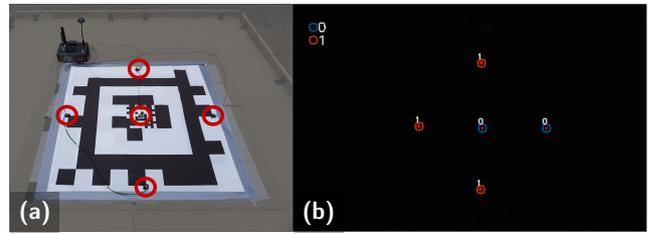}
  \caption{UV LEDs (marked with red circles) placed on the USV board (a) together with detected UV LEDs in the UV camera image (b).}
  \label{fig:landing_board_uvdar_pattern}
\end{figure}

The \ac{UVDAR} system was originally proposed for the mutual localization of \acp{UAV} in a swarm \citep{ahmad2021autonomous, novak2021predator, petracek2020swarms}.
Therefore, a novel pattern of \ac{UV} LEDs had to be defined in order to detect the horizontal landing board on the \ac{USV}.
We propose a 5-LED marker in the shape of a plus sign for the \ac{USV} board, as shown in \reffig{fig:landing_board_uvdar_pattern}.
Two LEDs (with 0 sign in \reffig{fig:landing_board_uvdar_pattern} (b)) got different signal codes than the others in order to identify the orientation of the marker unambiguously.
The data obtained from the \ac{UVDAR} system contain relative 3D position and orientation, which is used in the correction step of the Kalman filters (\refsec{sec:kalman_filters}).

\subsection{AprilTag detector} \label{sec:AprilTag_sensor}
\begin{figure}[!tb]
  \centering
  \input{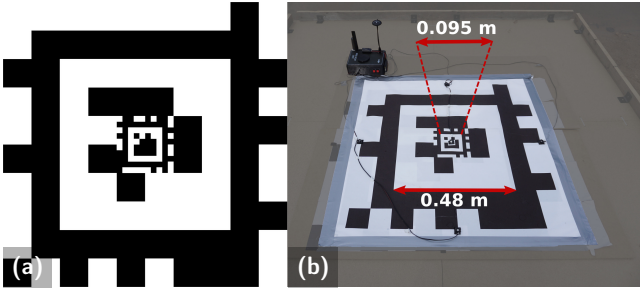}
  \caption{AprilTag layout (a) placed on the USV board (b).}
  \label{fig:apriltag_configuration}
\end{figure}

The AprilTag detector is a visual fiducial system that detects artificial landmarks known as AprilTags \citep{wang2016iros, olson2011tags}.
A single camera with a processing unit is sufficient to detect an AprilTag on a target.
The data measured by the AprilTag detector consist of the relative 3D position and orientation, which is used in the correction step of the Kalman filters (\refsec{sec:kalman_filters}).
The Apriltag detector supports custom tag layouts containing empty space in the middle for a recursive encapsulation of one tag in another \citep{Xu2020_vision, krogius2019iros}.
This feature improves the detectability from a wide range of distances by ensuring that a tag is fully visible even from a close distance.
The AprilTag layout placed on the \ac{USV} board is shown in \reffig{fig:apriltag_configuration}.
The custom tag layout is used, hence a smaller AprilTag is placed in the empty space of the bigger AprilTag.
The size of an AprilTag configuration on a real \ac{USV} board is shown in \reffig{fig:apriltag_configuration} (b).

Although the AprilTag detector provides less noisy data in close proximity than the \ac{UVDAR} system, the AprilTag detector requires sufficient light conditions for accurate AprilTag detection due to the passive markers.
The proposed setup combines the AprilTag detector and \ac{UVDAR} system to improve robustness and mainly show the possibility of fusing data from multiple sensors by the proposed approach.
In fact, almost any relevant sensory modality can be added to both proposed approaches, \usvesnonlinear{} and \usveslinear{}, if passing the innovation tests described in \refsec{sec:verification}.
As mentioned above, using these various detection approaches, we can verify the general usage of the proposed methodology, since almost any detection technique satisfying innovation tests described in \refsec{sec:verification} can be integrated into both proposed approaches, \usvesnonlinear{} and \usveslinear{}.

\section{VERIFICATION}
\label{sec:verification}
\begin{figure*}
  \centering
  \resizebox{1.0\textwidth}{!}{
   \input{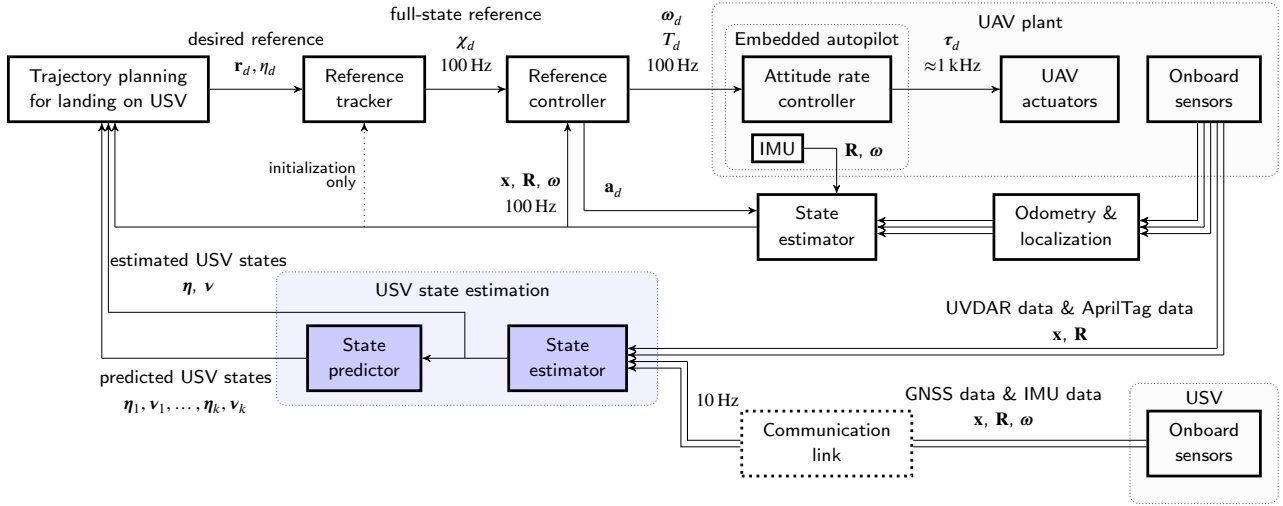}
   }
  \caption{
  Pipeline diagram of the proposed USV state estimation approach presented in this paper and integrated into the MRS system \citep{baca2021mrs} for experimental verification in realistic robotic scenarios. 
  \textit{Trajectory planning for landing on USV} block \citep{prochazka2024ModelPredictiveControlbased} computes the desired position and heading reference $\bm{(\mathbf{r}_d,~\eta_d)}$ for the UAV based on estimated and predicted USV states from the \textit{USV state estimation} block. 
  The computed trajectory is forwarded into the \textit{Reference tracker} that converts the trajectory to a smooth and feasible reference $\bm{\chi_d}$ for \textit{Reference controller}.
  The \textit{Reference controller} creates the desired thrust and angular velocities $\bm{(\omega_d,~T_d)}$ for the Pixhawk embedded flight controller.
  The \textit{State estimator} fuses data from \textit{Odometry \& localization} block to estimate the UAV translation, rotation, and angular velocities $\bm{(\mathbf{x},~\mathbf{R},~\bm{\omega})}$.
  The State estimator in \textit{USV state estimation} block fuses data from the UAV onboard sensors (UVDAR system and AprilTag detector) and USV onboard sensors (GNSS and IMU) to estimate the USV states $\bm{(\vec{\eta},~\vec{\nu})}$ using the Kalman filter.
  Then, the State predictor in \textit{USV state estimation} block predicts the future USV states $\bm{(\vec{\eta}_1,~\vec{\nu}_1,~\ldots,~\vec{\eta}_k,~\vec{\nu}_k)}$ based on the estimated USV states and a presented novel mathematical USV model.
  }
  \label{fig:pipeline_diagram}
\end{figure*}

The estimation approach presented in this paper was integrated into the \ac{MRS} \ac{UAV} system \citep{baca2021mrs}, as shown in \reffig{fig:pipeline_diagram}.
The USV state estimation block combines sensor data from both \ac{UAV} and \ac{USV} to estimate the current USV state and predict future states at 100~Hz.
The \ac{UAV} uses a trajectory generation method for agile landing presented in \citep{prochazka2024ModelPredictiveControlbased}, which leverages the estimated and predicted \ac{USV} states obtained from our approach presented in this paper.
The references computed by the trajectory planning module at 50~Hz for landing on the USV are processed by a Reference tracker to generate smooth and feasible full-state references at 100~Hz.
These are subsequently converted into thrust and angular velocity commands by the Reference controller.
The computed thrust and angular velocity commands are sent to the \ac{UAV}’s Pixhawk flight controller at 100~Hz.
Meanwhile, the \ac{UAV}’s state estimator continuously updates its translation, rotation, and linear and angular velocities at 100~Hz using data from the odometry and localization module.
Thereafter, the estimation approach was tested in the realistic Gazebo simulator extended by the \ac{VRX} simulator \citep{bingham19toward} (see \reffig{fig:uav_usv_landing_motivation} (d)). 
It was then verified by conducting real-world experiments.
The \ac{USV} used for verification weighs \SI{180}{kg} and measures \SI{5}{m} in length, \SI{2.5}{m} in width, and \SI{1.3}{m} in height.
The \ac{UAV} used for verification weighs \SI{3.5}{kg} and is \SI{0.15}{m} in height, with an arm length of \SI{0.325}{m}.
The extensive verification of the proposed approach performing state estimation is based on vector \ac{RMSE} \citep{rmse1}
and 3 innovation tests \citep{Evangelidis2021, kf_verification}.

Innovation is defined as
\begin{align}
    \vec{\zeta}(k) = \vec{y}(k) - \vec{\hat{y}}(k), \label{eq:innovation}
\end{align}
where $\vec{y}(k)$ is a measurement obtained from the sensors in time step $k$ and $\vec{\hat{y}}(k)$ is the expected measurement with respect to the current estimated state.
The covariance of innovation $ \vec{S}(k)$ has the following form:
\begin{align}
    \vec{S}(k) = \mathcal{E} \left\{  \vec{\zeta}(k)  \vec{\zeta}^{\intercal}(k) \right\},\label{eq:innovation_cov}
\end{align}
where $\mathcal{E}\{\cdot\}$ represents the mean value of a term inside the brackets.
If the filter is working correctly, then the mean of innovation $ \vec{\zeta}(k)$ is zero, and the matrix $\vec{S}(k)$ is its covariance matrix.

\subsection{Test 1: Innovation magnitude bound test}
The first test checks whether the innovation $ \vec{\zeta}(k)$ is consistent with its covariance $\vec{S}(k)$.
According to \citep{Evangelidis2021, kf_verification}, the innovation is consistent with its covariance if approximately $95\%$ of the innovation values are within the $95\%$ confidence ellipsoid defined by the matrix $\vec{S}(k)$.
In the scalar case, the test means that approximately $95\%$ of the innovation values lie within bounds $\pm 2 \sqrt{S(k)}$.

\subsection{Test 2: Normalized innovation squared $\chi^2$ test}
The second test seeks to prove the unbiasedness of the innovation \citep{Evangelidis2021, kf_verification}.
To perform the test, firstly, the normalized innovation squared $q(k)$ is determined as
\begin{align}
    q(k) = \vec{\zeta}^{\intercal}(k) \vec{S}^{-1}(k) \vec{\zeta}(k).
\end{align}
Then, the mean value of $q(k)$ is computed as
\begin{align}
   \bar{q} = \dfrac{1}{N} \sum_{k=1}^{N}  q(k). \label{eq:mean_normalized_innovation_squared}
\end{align}
To pass this test, the $\bar{q}$ should lie in confidence interval $[r_1,~r_2]$ characterized by the hypothesis $H_0$.
The hypothesis $H_0$ is defined as follows \citep{kf_verification}: $N\bar{q}$ is $\chi^2_{Nm}$ distributed with probability $P=1-\alpha$, where $m$ is a dimension of the measurement vector and $\alpha$ defines the confidence region, e.g., $\alpha=0.05$ specified 95\% confidence region,
\begin{align}
    P(N\bar{q} \in [r_1,~r_2]|H_0) = 1 - \alpha.\label{eq:test2_hypothesis_H0}
\end{align}

\subsection{Test 3: Innovation whiteness (autocorrelation) test}
The last test shows the whiteness of the innovation \citep{kf_verification}.
The time-averaged correlation is computed during the test 
\begin{align}
    corr(\tau) = \dfrac{1}{N} \sum_{k_r=0}^{N-\tau-1} \vec{\zeta}(k_r)^{\intercal} \vec{\zeta}(k_r+\tau),
\end{align}
which is normalized by $corr(0)$.
The idea of the test is that for a large enough $N$, $corr(\tau)$ is assumed to be normally distributed with zero mean and variance $\dfrac{1}{N}$ \citep{kf_verification}.
Therefore, at least $95\%$ of the values of $corr(\tau)$ should be in the confidence region defined as $\pm \dfrac{2}{\sqrt{N}}$.

\subsection{Verification of proposed approach \usvesnonlinear{}}
\begin{figure}[!t]
  \centering
  \includegraphics[width=\linewidth]{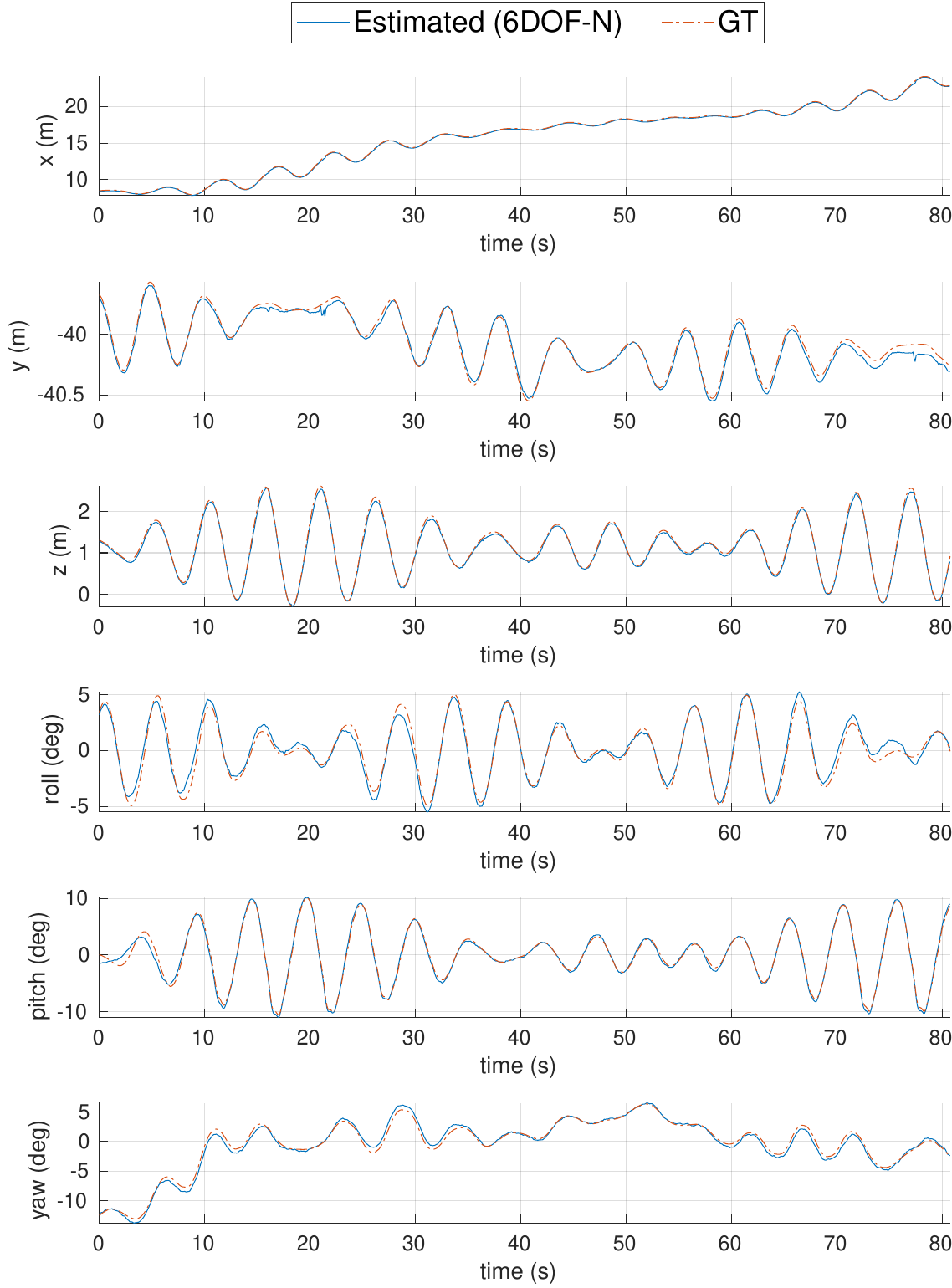} 
  \caption{Estimated position $(x,~y,~z)^{\intercal}$ and orientation $(\phi,~\theta,~\psi)^{\intercal}$ of the USV using the \usvesnonlinear{}.}
  \label{fig:ukf_estimations_pos_rot}
\end{figure}

\begin{figure}[!t]
  \centering
  \includegraphics[width=\linewidth]{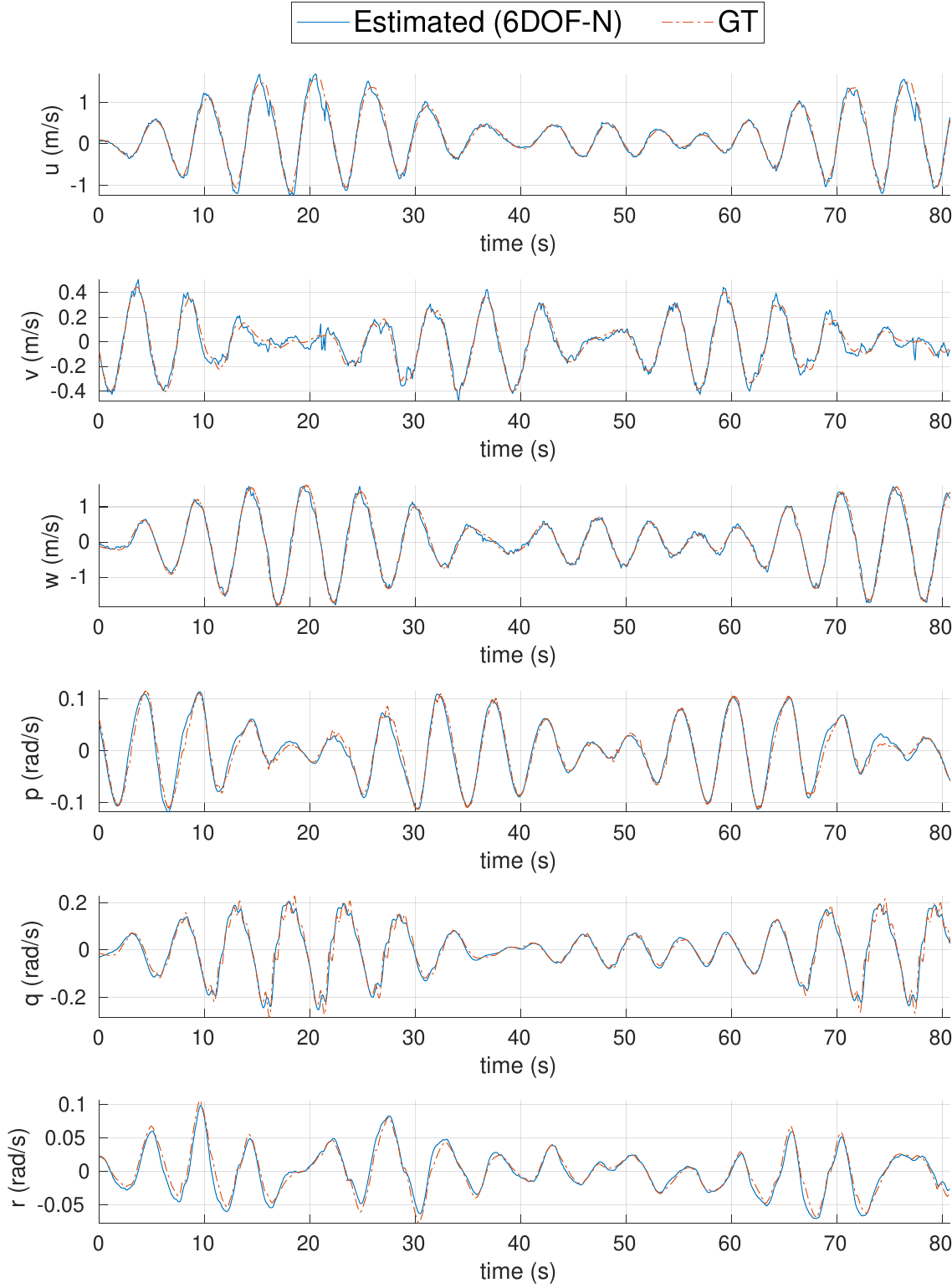} \caption{Estimated linear $(u,~v,~w)^{\intercal}$ and angular $(p,~q,~r)^{\intercal}$ velocities of the USV using the \usvesnonlinear{}.}
  \label{fig:ukf_estimations_lin_ang_vel}
\end{figure}

This section presents the verification of the proposed state estimation approach \usvesnonlinear{} (\refsec{sec:kalman_filters}).
The estimation of all \ac{USV} states using the \usvesnonlinear{} is presented in \reffig{fig:ukf_estimations_pos_rot} and \reffig{fig:ukf_estimations_lin_ang_vel}.
The estimated values nicely catch the \ac{GT} data as the \usvesnonlinear{} fuses data from all sensors.
The last row of \reftab{tab:UKF_RMSE} provides \ac{RMSE} of the estimated \ac{USV} states using the \usvesnonlinear{} computed from these experiments.

The \reftab{tab:UKF_innovation_tests} contains the results of the innovation tests applied to the \usvesnonlinear{} on \ac{USV} states $\vec{\eta}$ \refeq{eq:usv_global_states}, consisting of position $x,~y,~z$ and the angles roll $\phi$, pitch $\theta$, and yaw $\psi$.
The innovation tests are performed for individual sensors with respect to the \ac{USV} states that the sensors measure.
The results in \reftab{tab:UKF_innovation_tests} show that all sensors pass the innovation tests. 

\begin{table}[!b]
  \caption{RMSE of estimated USV states using the \usvesnonlinear{} according to the individual sensors.}
  \scriptsize
  \centering
    \begin{tabular}{lcccc}
    \hline
    sensor & \makecell{RMSE\\$(x,y,z)$\\m} & \makecell{RMSE\\$(\phi,\theta,\psi)$\\rad} & \makecell{RMSE\\$(u,v,w)$\\m/s} & \makecell{RMSE\\ $(p,q,r)$\\rad/s}\\
    \hline
    GPS & 0.681 & - & 0.837 & - \\
    IMU & - & 0.011 & - & 0.087 \\
    UVDAR & 0.289 & 0.123 & 0.385 & 0.248 \\
    AprilTag & 0.046 & 0.061 & 0.146 & 0.092 \\
    all sensors & 0.097  & 0.016  & 0.157 & 0.025 \\
    \hline
    \end{tabular}
  \label{tab:UKF_RMSE}
\end{table}

\begin{table}[!b]
  \caption{Innovation tests applied to the \usvesnonlinear{} according to the individual sensors.}
  \scriptsize
  \centering
    \begin{tabular}{lcccc}
    \hline
    sensor & states & test 1 & \makecell{test 2\\($q$ of $[q_{min},~q_{max}]$)} & test 3 \\ 
    \hline
    GPS & $(x,y,z)$ &  98.9\% & 2162.2 of [1998.2, 2253.6] & 99.8\% \\
    IMU & $(\phi,\theta,\psi)$ & 93.1\% & 1040.7 of [893.2, 1066.6] & 94.8\%\\
    UVDAR & $(x,y,z)$ & 96.4\% & 571.3 [547.2, 684.6] & 95.8\% \\
    UVDAR & $(\phi,\theta,\psi)$ & 98.4\% & 590.7 of [547.2, 684.6] & 99.0\%\\
    AprilTag & $(x,y,z)$ & 99.1\% & 8568.1 of [8510.4, 9029.4] & 95.3\% \\
    AprilTag & $(\phi,\theta,\psi)$ & 93.6\% & 8725.6 of [8510.3, 9029.4] & 97.7\%\\
    \hline
    \end{tabular}
  \label{tab:UKF_innovation_tests}
\end{table}

The \ac{RMSE} according to the individual sensors used to estimate the desired \ac{USV} states is proposed in \reftab{tab:UKF_RMSE}.
The largest \ac{RMSE} from all sensors for states $(x,~y,~z)^{\intercal}$ and $(u,~v,~w)^{\intercal}$ is provided by estimation using the \ac{GPS} sensor.
In contrast, the estimation of the states $(\phi,~\theta,~\psi)^{\intercal}$ and $(p,~q,~r)^{\intercal}$ using the \ac{IMU} sensor achieve the best results from all the sensors as the values of \ac{IMU} \ac{RMSE} for states $(\phi,~\theta,~\psi)^{\intercal}$ and $(p,~q,~r)^{\intercal}$ are the lowest.
The estimated \ac{USV} states $(x,~y,~z)^{\intercal}$ and $(u,~v,~w)^{\intercal}$ that use the \ac{UVDAR} system have a twice smaller \ac{RMSE} than states estimated by the \ac{GPS}.
However, the states $(\phi,~\theta,~\psi)^{\intercal}$ estimated using the \ac{UVDAR} system have a ten times greater \ac{RMSE} than estimations using the \ac{IMU}.
The \ac{RMSE} for states $(p,~q,~r)^{\intercal}$ is three times smaller for the \ac{IMU} in comparison with the \ac{UVDAR}.
The smallest \ac{RMSE} for states $(x,~y,~z)^{\intercal}$ and $(u,~v,~w)^{\intercal}$ from all sensors is achieved using the AprilTag.
However, the \ac{RMSE} of the AprilTag for $(\phi,~\theta,~\psi)^{\intercal}$ is five times higher than  the \ac{RMSE} of the \ac{IMU}. 
The \ac{RMSE} for states $(p,~q,~r)^{\intercal}$ using the AprilTag is slightly higher than \ac{RMSE} for these states using the \ac{IMU}.

The estimated states using the \usvesnonlinear{} are used to predict future \ac{USV} states applying the prediction steps as described in \refsec{sec:kalman_filters}.
Every two seconds, the \ac{USV} states are predicted for two seconds. 
The predicted \ac{USV} states $(x,~y,~z,~\phi,~\theta,~\psi)^{\intercal}$ are shown in \reffig{fig:ukf_predictions}.
The graphs in \reffig{fig:ukf_predictions} demonstrate that the \usvesnonlinear{} is able to estimate and predict the motion of the \ac{USV}.

The \ac{RMSE} of states $(x,~y,~z)^{\intercal}$ is six times larger for predicted states than for estimated ones (\reftab{tab:ukf_rmse_prediction}). 
However, the \ac{RMSE} of predicted states $(x,~y,~z)^{\intercal}$ corresponds to the \ac{RMSE} of estimated states $(x,~y,~z)^{\intercal}$ using only the \ac{GPS} sensor (\reftab{tab:UKF_RMSE}).
The \ac{RMSE} of states $(\phi,~\theta,~\psi)^{\intercal}$ is five times larger for predicted states than for estimated ones (\reftab{tab:ukf_rmse_prediction}). 
The \ac{RMSE} of predicted states $(\phi,~\theta,~\psi)^{\intercal}$ is smaller than \ac{RMSE} of estimated states $(\phi,~\theta,~\psi)^{\intercal}$ using the \ac{UVDAR} (\reftab{tab:UKF_RMSE}).
However, \ac{RMSE} of predicted states $(\phi,~\theta,~\psi)^{\intercal}$ is still larger than the \ac{RMSE} of the states estimated by the AprilTag detector (\reftab{tab:UKF_RMSE}).

\begin{table}[!bt]
  \caption{RMSE of predicted and estimated USV states using the \usvesnonlinear{}.}
  \scriptsize
  \centering
    \begin{tabular}{lccc}
    \hline
    \ac{USV} states & \makecell{RMSE\\$(x,~y,~z)$\\m} & \makecell{RMSE\\$(\phi,~\theta,~\psi)$\\rad}\\ 
    \hline
    predicted states & 0.647 & 0.081\\
    estimated states & 0.097 & 0.016\\
    \hline
    \end{tabular}
    \label{tab:ukf_rmse_prediction}
\end{table}

\begin{figure}[!t]
  \centering
  \input{figure10.tex}
  \caption{Predicted and estimated position $(x,~y,~z)^{\intercal}$ and orientation $(\phi,~\theta,~\psi)^{\intercal}$ of the USV using the \usvesnonlinear{}. The rectangles highlight zoomed-in regions of the graphs.}
  \label{fig:ukf_predictions}
\end{figure}

\subsection{Comparison of \usvesnonlinear{}, \usveslinear{} and state-of-the-art methods}
\label{sec:comparison_lkf_and_ukf}

First, we compare the two proposed methods, \usvesnonlinear{} and \usveslinear{} (\refsec{sec:kalman_filters}).
Then, we further compare both methods with the state-of-the-art.
The main difference between the proposed methods is in the \ac{USV} model: the \usvesnonlinear{} uses the nonlinear \ac{USV} model extended by the nonlinear wave model (\refsec{sec:usv_nonlinear_model} and \ref{sec:waves_model_nonlin}), while the \usveslinear{} uses the linear \ac{USV} model obtained by linearization of the nonlinear \ac{USV} model (\refsec{sec:usv_linear_model}).
The linear \ac{USV} model is extended by the linear wave model (\refsec{sec:waves_model_lin}).
The \usveslinear{} benefits from using \ac{LKF} as it is less computationally demanding than \usvesnonlinear{} that applies \ac{UKF} on full nonlinear \ac{USV} model.
From our tests performed on real \ac{UAV}, we found that \usveslinear{} is 30\% faster than \usvesnonlinear{}.

The innovation tests applied on the \usvesnonlinear{} are also performed for the \usveslinear{} (\reftab{tab:LKF_innovation_tests}).
The innovation tests demonstrate that estimations using the \usveslinear{} are consistent with the sensors' measurements.
However, test 1 of the \ac{UVDAR} system for states $(x,~y,~z)^{\intercal}$ reaches a value 100\%, but this value should be near 95\%.
The sensor covariance matrix $\vec{R}_{uvdar}$ is over-estimated in the \usveslinear{} for states $(x,~y,~z)^{\intercal}$ measured by the \ac{UVDAR} system.
The reason involves the high peaks in \ac{UVDAR} measurements, which would negatively impact estimation.
The innovation tests of the \usvesnonlinear{} prove that estimations using the \usvesnonlinear{} are consistent with the sensors' measurements, even for the \ac{UVDAR} system.

The \ac{RMSE} of predicted states $(x,~y,~z)^{\intercal}$ and $(\phi,~\theta,~\psi)^{\intercal}$ using \usveslinear{} is larger than \ac{RMSE} of \usvesnonlinear{} (\reftab{tab:lkf_rmse_prediction}).
The \usveslinear{} has a 14\% larger \ac{RMSE} for states $(x,~y,~z)^{\intercal}$ compared to the \usvesnonlinear{}.
The \ac{RMSE} of predicted states $(\phi,~\theta,~\psi)^{\intercal}$ using \usvesnonlinear{} is two and a half times smaller than when using \usveslinear{}.

The \ac{RMSE} of estimated states $(u,~v,~w)^{\intercal}$, $(p,~q,~r)^{\intercal}$ and $(\phi,~\theta,~\psi)^{\intercal}$ is smaller for \usvesnonlinear{} (\reftab{tab:sota_RMSE}).
However, the \ac{RMSE} of estimated states $(x,~y,~z)^{\intercal}$ is slightly smaller for \usveslinear{}, mainly caused by the higher computational demands of \usvesnonlinear{} compared to \usveslinear{}.
As mentioned above, the \usveslinear{} is 30\% faster than \usvesnonlinear{}.
Therefore, \usveslinear{} is less delayed and enables faster control loop that results in faster \ac{UAV} landing and longer mission time, as spending less time on landing increases the \ac{UAV} flight time.
Nevertheless, for all other states, the \usvesnonlinear{} provides a more accurate estimation of the \ac{USV} states because the nonlinear \ac{USV} model better captures the \ac{USV} motion dynamics.

\begin{table}[!t]
    \caption{Innovation tests applied to the \usveslinear{} according to the individual sensors.}
   \scriptsize
   \centering
   \begin{tabular}{lcccc}
     \hline
     sensor & states & test 1 & \makecell{test 2\\($q$ of $[q_{min},~q_{max}]$)} & test 3 \\ 
     \hline
     GPS & $(x,y,z)$ &  98.2\% & 644.7 of [590.7, 733.1] & 99.7\% \\
     IMU & $(\phi,\theta,\psi)$ & 93.7\% & 1008.7 of [866.5, 1037.3] & 98.5\%\\
     UVDAR & $(x,y,z)$ &  100\% & 683.6 of [824.5, 991.3] & 96.9\% \\
     UVDAR & $(\phi,\theta,\psi)$ & 96.5\% & 830.4 of [824.5, 991.3] & 99.3\%\\
     AprilTag & $(x,y,z)$ & 96.7\% &  8701.4 of [8486.7, 9005.1] & 97.3\% \\
     AprilTag & $(\phi,\theta,\psi)$ & 95.5\% & 8800.2 of [8486.7, 9005.1] & 98.5\%\\
     \hline
     \end{tabular}
   \label{tab:LKF_innovation_tests}
\end{table}

\begin{table}[!t]
\caption{RMSE of predicted USV states using the \usvesnonlinear{} and \usveslinear{}.}
\scriptsize
\centering
 \begin{tabular}{lccc}
 \hline
 predicted \ac{USV} states & \makecell{RMSE\\$(x,~y,~z)$\\m} & \makecell{RMSE\\$(\phi,~\theta,~\psi)$\\rad}\\ 
 \hline
    \usvesnonlinear{} & \textbf{0.647} & \textbf{0.081}\\
    \usveslinear{} & 0.737 & 0.196\\
 \hline
 \end{tabular}
\label{tab:lkf_rmse_prediction}
\end{table}

In order to compare our approach with state-of-the-art methods presented in \refsec{sec:related_works}, we firstly consider works \citep{wang2021_dual_ukf, Xu2020_vision, Wirtensohn2016518, tomera2012nonlinear, KalmanFossen} that are able to estimate the \ac{USV} only with 3 \acp{DOF}, i.e. $x$, $y$ and yaw angle.
Therefore no information about $z$ (heave) position and roll and pitch angles are provided. 
However, these states have a significant influence on \ac{USV} motion relatively to the \ac{UAV} as seen in \reffig{fig:ukf_estimations_pos_rot} and \reffig{fig:ukf_estimations_lin_ang_vel}.
Not considering these states during autonomous landing can lead to unsafe maneuvers or crashes, as shown in \reffig{fig:uav_usv_landing_motivation} (a) where a heave motion of a ship damages a landing helicopter.
We observed the same problems in our initial simulations, and due to the high rise of collision, we could not repeat the evaluation in real-world conditions using these 3 \ac{DOF} approaches.

\begin{figure}[!t]
  \centering
  \includegraphics[width=\linewidth]{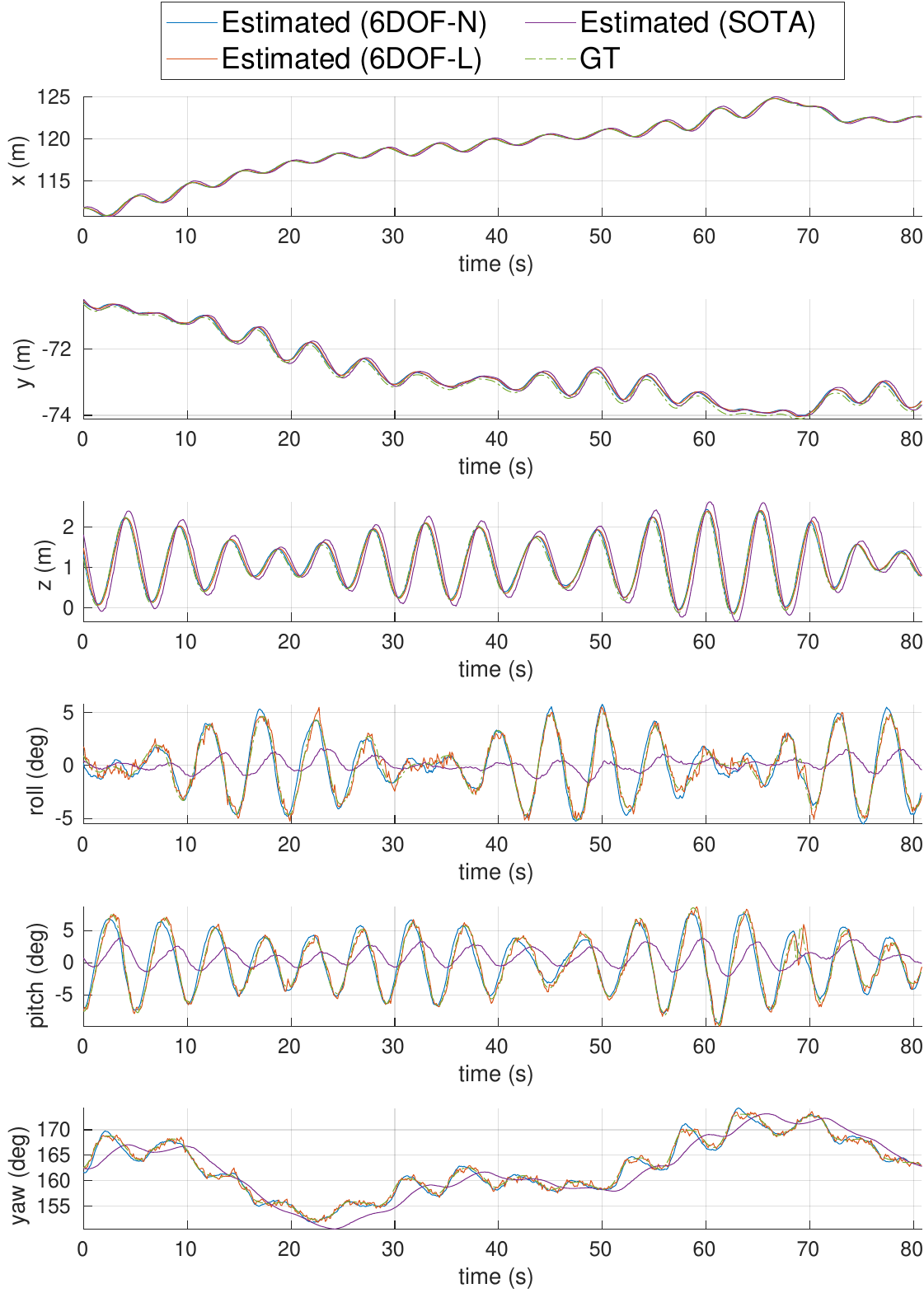} 
  \caption{Estimated position $(x,~y,~z)^{\intercal}$ and orientation $(\phi,~\theta,~\psi)^{\intercal}$ of the USV using the \usvesnonlinear{}, \usveslinear{} and state-of-the-art method (\sota{}).}
  \label{fig:sota_estimations_pos_rot}
\end{figure}

\begin{table}[!b]
  \caption{RMSE of estimated USV states using the \usvesnonlinear{}, \usveslinear{} and state-of-the-art method (\sota{}).}
  \scriptsize
  \centering
    \begin{tabular}{lccccc}
    \hline
    sensor & \makecell{RMSE\\$(x,y,z)$\\m} & \makecell{RMSE\\$(\phi,\theta,\psi)$\\rad} & \makecell{RMSE\\$(u,v,w)$\\m/s} & \makecell{RMSE\\$(p,q,r)$\\rad/s}\\ 
    \hline
    \usvesnonlinear{} & 0.120 & \textbf{0.013} & \textbf{0.201} & \textbf{0.204}\\
    \usveslinear{} & \textbf{0.116} & 0.017 & 0.303 & 0.235\\
    \sota{} & 0.313 & 0.056 & 0.580 & 0.329\\
    \hline
    \end{tabular}
  \label{tab:sota_RMSE}
\end{table}

Therefore, we present a comparison of \ac{USV} state estimation with 6 \acp{DOF} between proposed approaches \usvesnonlinear{} and \usveslinear{} and state-of-the-art method \citep{uav_usv_landing2}, which we call \sota{}.
The approach \citep{Abujoub2018_landing} is very similar to the \sota{} method \citep{uav_usv_landing2}, however, the \citep{Abujoub2018_landing} uses only 5 \acp{DOF}.
The \reffig{fig:sota_estimations_pos_rot} shows estimated position $(x,~y,~z)^{\intercal}$ and orientation $(\phi,~\theta,~\psi)^{\intercal}$.
The \sota{} had a non-negligible problem in estimating wave motions in orientation and also incorrectly estimated a larger amplitude in state $z$ together with a larger delay than the proposed approaches \usvesnonlinear{} and \usveslinear{}.
The better performance of proposed approaches \usvesnonlinear{} and \usveslinear{} compared to \sota{} is demonstrated by the \ac{RMSE} values presented in \reftab{tab:sota_RMSE}.
The \ac{RMSE} of \sota{} is the largest for all \ac{USV} states $(x,~y,~z)^{\intercal}$, $(\phi,~\theta,~\psi)^{\intercal}$, $(u,~v,~w)^{\intercal}$, and $(p,~q,~r)^{\intercal}$ compared to the \usvesnonlinear{} and \usveslinear{}.
The biggest difference is in estimation of orientation $(\phi,~\theta,~\psi)^{\intercal}$, as the \sota{} has the \ac{RMSE} three times larger than \usvesnonlinear{} and \usveslinear{}.

To summarize, our method offers accurate \ac{USV} state estimation and prediction of future \ac{USV} states on a wavy water surface, enabled by the novel \ac{USV} mathematical model that incorporates wave dynamics.
Our approach provides estimations and predictions in all 6 \acp{DOF}, which is essential for tasks like \ac{UAV} landing on a \ac{USV} moving on a wavy water surface.
By fusing data from multiple \ac{UAV} and \ac{USV} sensors, our approach ensures robustness and mitigates risks associated with sensor malfunctions.
Furthermore, our method provides covariance for each estimated and predicted USV state, quantifying the accuracy of these estimations and predictions.
However, our method also has limitations that warrant future research.
One of the most challenging situations for the presented approach arises when the \ac{USV} performs aggressive turns during motion.
In such cases, the predictions may become inaccurate due to spikes in \ac{USV} velocity, requiring some time to converge again.
Additionally, the method's ability to adapt in real-time to rapidly changing wave conditions is limited.
As shown in \reffig{fig:ukf_predictions}, the accuracy of state predictions improves over time.
When wave characteristics change rapidly, our method must adapt to these new conditions.


\section{REAL-WORLD EXPERIMENTS}
After a successful verification in the realistic robotic simulator Gazebo, the estimation approach presented in this paper was deployed in real-world experiments.
Multimedia materials supporting the results of this paper are available at
\paperlink{}.
Numerous real-world experiments were performed, with two of them representing two different application scenarios presented in this section.
Firstly, the \ac{UAV} followed the moving \ac{USV} with artificially-induced wave motion.
Secondly, the \ac{UAV} took off and flew above the \ac{USV}.
Then, the \ac{UAV} followed the moving \ac{USV} using estimated and predicted states of the \ac{USV}.
During the \ac{USV} following, the \ac{UAV} used the predicted \ac{USV} states to select the moment for landing.
Finally, the \ac{UAV} landed on the estimated landing platform located on the \ac{USV}. 

\begin{figure}[!b]
  \centering
  \input{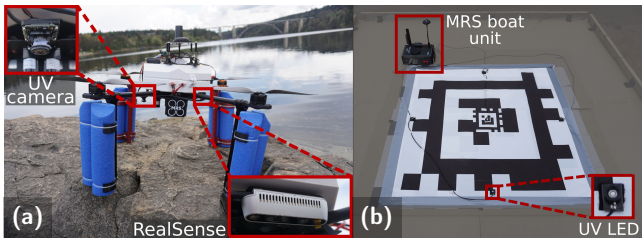}
  \caption{The UAV (a) and USV landing platform (b) used in real-world experiments.}
  \label{fig:real_uav_usv}
\end{figure}

\begin{figure}[!bt]
    \centering    
    \input{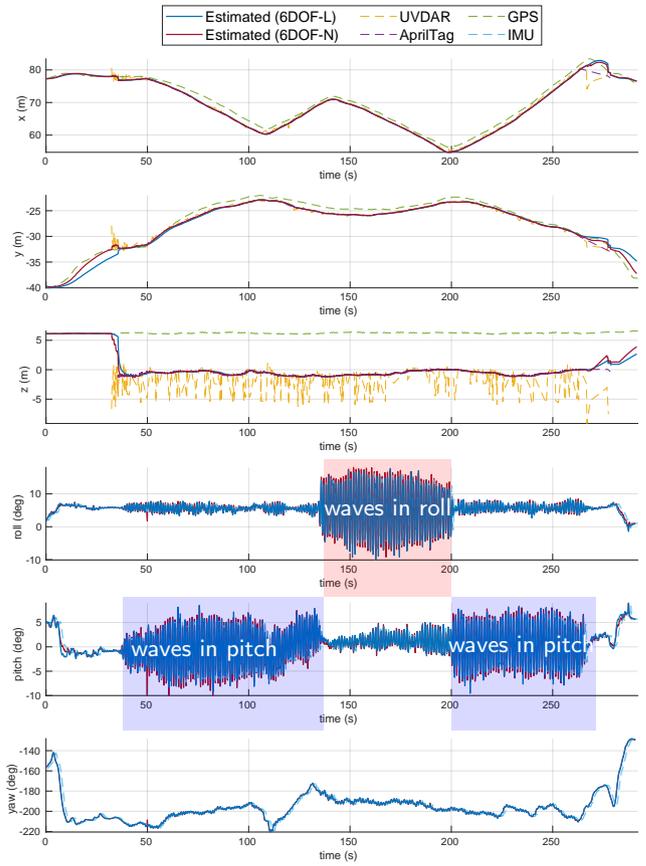}
    \vspace{-5mm}
    \caption{Estimation of the USV states $(x,~y,~z,~\phi,~\theta,~\psi)^{\intercal}$ obtained using the \usveslinear{} and the \usvesnonlinear{} while the UAV was following the moving and oscillating USV.}
    \label{fig:following_boat_estimations}
\end{figure}

The \ac{UAV} used in real-world experiments (\reffig{fig:real_uav_usv} (a)) is built on a Tarot T650 frame, equipped with brushless direct current Tarot 4114 320KV motors and Turnigy Multistar 51A electronic speed controllers.
The \ac{UAV} is powered by a LiPo 6S \SI{8000}{mAh} battery and uses a Pixhawk~4 autopilot with \ac{GPS}.
The overall \ac{UAV} autonomous system, including our approach, runs onboard the \ac{UAV} using a NUC8i7BEH high-level computer equipped with an i7-8559U processor, \SI{16}{GB} of DDR4 RAM, and a \SI{500}{GB} SSD (see \citep{HertJINTHW_paper, MRS2022ICUAS_HW} for details about our hardware equipment).
The~\ac{UAV} is also equipped with the necessary sensors for the estimation of \ac{USV} states --- the RealSense D435 camera for the AprilTag detector and the mvBlueFOX MLC200wG
camera with \ac{UV} bandpass filter for the \ac{UVDAR} system.
All electronic components are protected with a water-resistant spray and placed in a 3D-printed waterproof cover.

The \reffig{fig:real_uav_usv} (b) shows the \SI{2}{m}~$\times$~\SI{2}{m} square landing platform placed on the \ac{USV} during real-world experiments.
The landing spot is defined by an AprilTag (\SI{0.8}{m}~$\times$~\SI{0.8}{m}) centered on the \ac{USV} board (\refsec{sec:AprilTag_sensor}).
The four \ac{UV} LEDs are located around the perimeter of the AprilTag, and one \ac{UV} LED is placed in the center of the AprilTag (\refsec{sec:UVDAR_sensor}).
The \ac{MRS} boat unit is installed next to the AprilTag and contains a NUC8i7BEH computer, LiPo 4S \SI{6750}{mAh} battery, \ac{GPS} module, and \ac{IMU} sensor (\refsec{sec:sensors_placed_on_USV}).

In the first real-world scenario, the \ac{UAV} followed the moving \ac{USV} (see video\footnote{\paperlink{}\label{footnote:mrs_page_novak}}).
The \reffig{fig:following_boat_estimations} presents the estimated \ac{USV} states $x,~y,~z$, roll $\phi$, pitch $\theta$, and yaw $\psi$ in one of the experimental flights using the proposed approaches \usveslinear{} and \usvesnonlinear{} (\refsec{sec:kalman_filters}).
The \ac{UAV} onboard sensors provided the first data at time $t=30$~s that improved the estimation of the \ac{USV} states, especially in the $z$ position.
The~graphs of roll $\phi$ and pitch $\theta$ contain wave motions marked in the \reffig{fig:following_boat_estimations}.
Both approaches \usveslinear{} and \usvesnonlinear{} captured these motions in their estimations.
The graphs of roll, pitch, and yaw do not contain the \ac{UVDAR} and AprilTag measurements as these measurements are very noisy.

\begin{figure}[!t]
    \centering    
    \input{figure14.tex}
    \vspace{-5mm}
    \caption{Estimation of the USV states $(x,~y,~z,~\phi,~\theta,~\psi)^{\intercal}$ using the \usveslinear{} and \usvesnonlinear{} while the UAV followed the USV and then landed on it. The rectangles highlight zoomed-in regions of the graphs.}
    \label{fig:land_slow_moving_boat_estimations}
\end{figure}

\begin{figure}[!tb]
  \centering
  \input{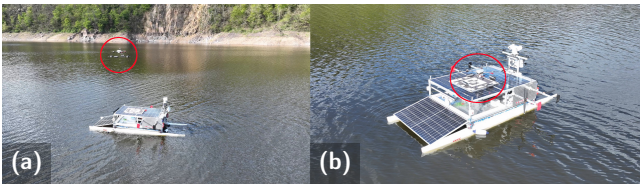}
  \caption{Snapshots from one of the real-world experiments in which the UAV followed the USV (a). The experiment was finished with a~successful landing (b).}
  \label{fig:snapshots_rw_uav_follow_usv_and_land_on_it}
\end{figure}

In the second scenario, the \ac{UAV} flew above the \ac{USV} and estimated the \ac{USV} states using only the data received from the \ac{USV} onboard sensors --- the \ac{GPS} and the \ac{IMU}.
We provide data from one of the experimental flights in \reffig{fig:land_slow_moving_boat_estimations}.
After the \ac{UAV} onboard sensors (\ac{UVDAR} system and AprilTag detector) detected the desired markers in their camera frames, the measurements from these \ac{UAV} onboard sensors improved the estimations of the \ac{USV} states.
The \ac{UAV} followed the moving \ac{USV} for 98~s using the estimated and predicted \ac{USV} states.
While following the \ac{USV}, the \ac{UAV} was looking for acceptable conditions to land on the \ac{USV} and canceled the landing maneuver whenever the landing was unsafe for the \ac{UAV}.
At time $t=142$~s, the \ac{UAV} successfully landed on the \ac{USV} (see video\textsuperscript{\ref{footnote:mrs_page_novak}}).

\begin{figure}[!t]
  \centering
  \includegraphics[width=\linewidth]{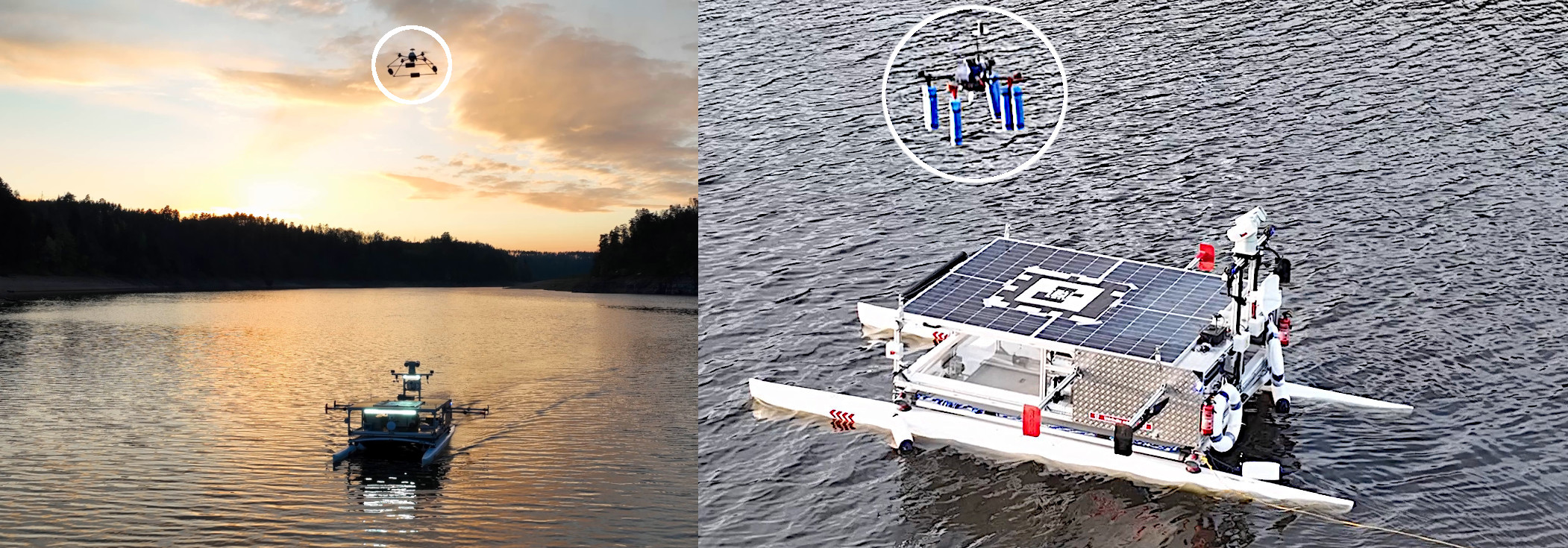}
  \caption{Snapshot from testing of the proposed approach integrated into a complex system of water-proofed floatable \ac{UAV} and fully autonomous \ac{USV} cooperating in various real-world conditions.}
  \label{fig:uav_usv_hitech}
\end{figure}

The estimations of the \ac{USV} states from the proposed approaches \usveslinear{} and \usvesnonlinear{} during this real-world experiment are presented in \reffig{fig:land_slow_moving_boat_estimations}.
The \ac{UAV} onboard sensors began providing data at time $t=45$~s and improved estimation of the \ac{USV} states; this is prevalent especially in the graph of the \ac{USV} state $z$.
The \ac{UVDAR} measurements of $z$, roll $\phi$, pitch $\theta$, yaw $\psi$, and the AprilTag measurements of roll $\psi$ and pitch $\theta$ were very noisy.
However, the \usveslinear{} and \usvesnonlinear{} filtered the noise out and were able to provide a smooth estimation of the \ac{USV} states.
The snapshots from one of these real-world experiments are shown in \reffig{fig:snapshots_rw_uav_follow_usv_and_land_on_it}.
The reliability and practical usefulness of the proposed approach were also demonstrated during the testing of a high-tech system of cooperating \ac{UAV}-\ac{USV} designed for water quality monitoring and garbage removal (see \reffig{fig:uav_usv_hitech}).

\section{CONCLUSION}
This paper presents a novel approach for the state estimation and prediction of a \ac{USV} moving on a rough water surface.
This work introduces a novel nonlinear model of the \ac{USV} in 6 \acp{DOF} extended by wave motion dynamics tailored for \ac{UAV}-\ac{USV} tight collaboration.
The usage of the \ac{USV} mathematical model allows for estimating all 6 \ac{DOF} states of the \ac{USV} moving in waves, and also predicting the future \ac{USV} states including wave motions.
In order to achieve the required reliability, the estimation pipeline is able to fuse data from multiple sensors. 
Thus, the presented approach is capable of being used in different real-world conditions and at varying distances between the \ac{UAV} and the \ac{USV}.
The performance of estimation and prediction was exhaustively verified in simulations and was compared with state-of-the-art methods.
The comparison with the state-of-the-art methods demonstrated that the proposed approach significantly surpassed these methods.
Finally, the approach integrated into a complex \ac{UAV}-\ac{USV} system was deployed and experimentally verified in real-world conditions, where the \ac{UAV} was successfully following the \ac{USV} and repeatably landing on its landing deck.

\bibliographystyle{cas-model2-names}
\bibliography{main}

\begin{acronym}
  \acro{CNN}[CNN]{Convolutional Neural Network}
  \acro{IR}[IR]{infrared}
  \acro{GNSS}[GNSS]{Global Navigation Satellite System}
  \acro{MOCAP}[mo-cap]{Motion capture}
  \acro{MPC}[MPC]{Model Predictive Control}
  \acro{MRS}[MRS]{Multi-robot Systems group}
  \acro{ML}[ML]{Machine Learning}
  \acro{MAV}[MAV]{Micro-scale Unmanned Aerial Vehicle}
  \acro{UAV}[UAV]{Unmanned Aerial Vehicle}
  \acro{UV}[UV]{ultraviolet}
  \acro{UVDAR}[\emph{UVDAR}]{UltraViolet Direction And Ranging}
  \acro{UT}[UT]{Unscented Transform}
  \acro{RTK}[RTK]{Real-Time Kinematic}
  \acro{ROS}[ROS]{Robot Operating System}
  \acro{wrt}[w.r.t.]{with respect to}
  \acro{LTI}[LTI]{Linear time-invariant}
  \acro{USV}[USV]{Unmanned Surface Vehicle}
  \acroplural{DOF}[DOFs]{Degrees of Freedom}
  \acro{DOF}[DOF]{Degree of Freedom}
  \acro{API}[API]{Application Programming Interface}
  \acro{CTU}[CTU]{Czech Technical University}
  \acroplural{DOF}[DOFs]{Degrees of Freedom}
  \acro{DOF}[DOF]{Degree of Freedom}
  \acro{FOV}[FOV]{Field of View}
  \acro{GNSS}[GNSS]{Global Navigation Satellite System}
  \acro{GPS}[GPS]{Global Positioning System}
  \acro{IMU}[IMU]{Inertial Measurement Unit}
  \acro{LKF}[LKF]{Linear Kalman Filter}
  \acro{LTI}[LTI]{Linear time-invariant}
  \acro{LiDAR}[LiDAR]{Light Detection and Ranging}
  \acro{MAV}[MAV]{Micro Aerial Vehicle}
  \acro{MPC}[MPC]{Model Predictive Control}
  \acro{MRS}[MRS]{Multi-robot Systems}
  \acro{ROS}[ROS]{Robot Operating System}
  \acro{RTK}[RTK]{Real-time Kinematics}
  \acro{SLAM}[SLAM]{Simultaneous Localization And Mapping}
  \acro{UAV}[UAV]{Unmanned Aerial Vehicle}
  \acro{UGV}[UGV]{Unmanned Ground Vehicle}
  \acro{UKF}[UKF]{Unscented Kalman Filter}
  \acro{USV}[USV]{Unmanned Surface Vehicle}
  \acro{RMSE}[RMSE]{Root Mean Square Error}
  \acro{UVDAR}[UVDAR]{UltraViolet Direction And Ranging}
  \acro{UV}[UV]{UltraViolet}
  \acro{VRX}[VRX]{Virtual RobotX}
  \acro{WAM-V}[WAM-V]{Wave Adaptive Modular Vessel}
  \acro{GT}[GT]{Ground Truth}
  \acro{UTM}[UTM]{Universal Transverse Mercator}
\end{acronym}

\end{document}